\documentclass[twoside,leqno,twocolumn]{article}

\usepackage[letterpaper]{geometry}
\usepackage{adjustbox}
\usepackage{ltexpprt}
\usepackage{hyperref}
\usepackage{amsmath,amsfonts}
\usepackage{mathrsfs}
\usepackage{url}

\usepackage{amssymb}
\usepackage{subfigure}
\usepackage[justification=centering]{caption}
\usepackage{soul}
\usepackage{xcolor}
\usepackage{booktabs}
\usepackage{algorithm}
\usepackage{algorithmic}

\begin{document}

\newcommand\relatedversion{}

\title{\textbf{Traceable Automatic Feature Transformation via Cascading Actor-Critic Agents}}
\author{Meng Xiao$^{1,2}$ \and Dongjie Wang$^{3}$ \and Min Wu$^4$ \and Ziyue Qiao$^5$ \and Pengfei Wang$^1$  \and Kunpeng Liu$^6$ \and Yuanchun Zhou$^{1,*}$ \and Yanjie Fu$^{3,*}$}
\date{} 
\maketitle
\fancyfoot[R]{\scriptsize{Copyright \textcopyright\ 2023 by SIAM\\
Unauthorized reproduction of this article is prohibited}}
\let\thefootnote\relax\footnotetext{{\scriptsize $^1$ Computer Network Information Center, CAS, Emails:\\ shaow@cnic.cn, wpf2106@gmail.com, zyc@cnic.cn}}
\let\thefootnote\relax\footnotetext{\scriptsize $^2$ University of Chinese Academy of Sciences}
\let\thefootnote\relax\footnotetext{\scriptsize $^3$ University of Central Florida, Emails:\\ wangdongjie@knights.ucf.edu, yanjie.fu@ucf.edu}
\let\thefootnote\relax\footnotetext{\scriptsize $^4$ Institute for Infocomm Research, A*STAR, E-mail:\\ wumin@i2r.a-star.edu.sg}
\let\thefootnote\relax\footnotetext{\scriptsize $^5$ The Hong Kong University of Science and Technology (Guangzhou), Email: zyqiao@ust.hk}
\let\thefootnote\relax\footnotetext{\scriptsize $^6$ Portland State University, Email: kunpeng@pdx.edu}
\let\thefootnote\relax\footnotetext{\scriptsize $^*$  Corresponding authors}


\begin{abstract} 
\small\baselineskip=9pt
Feature transformation for AI is an essential task to boost the effectiveness and interpretability of machine learning (ML). Feature transformation aims to transform original data to identify an optimal feature space that enhances the performances of a downstream ML model. Existing studies either combines preprocessing, feature selection, and generation skills to empirically transform data,  or automate feature transformation by machine intelligence, such as reinforcement learning. However, existing studies suffer from: 1) high-dimensional non-discriminative feature space; 2) inability to represent complex situational states;  3) inefficiency in integrating local and global feature information. To fill the research gap, we propose a novel group-wise cascading actor-critic perspective to develop the AI construct of automated feature transformation. 
Specifically, we formulate the feature transformation task as an iterative, nested process of feature generation and selection, where feature generation is to generate and add new features based on original features, and feature selection is to remove redundant features to control the size of feature space. Our proposed framework has three technical aims: 1) efficient generation; 2) effective policy learning; 3) accurate state perception. For an efficient generation, we develop a tailored feature clustering algorithm and accelerate generation by feature group-group crossing based generation. For effective policy learning, we propose a cascading actor-critic learning strategy to learn state-passing agents to select candidate feature groups and operations for fast feature generation. Such a strategy can effectively learn policies when the original feature size is large, along with exponentially growing feature generation action space, in which classic Q-value estimation methods fail. For accurate state perception of feature space, we develop a state comprehension method considering not only pointwise feature information but also pairwise feature-feature correlations. Finally, we present extensive experiments and case studies to illustrate 24.7\% improvements in F1 scores compared with SOTAs and robustness in high-dimensional data.\let\thefootnote\relax\footnotetext{the release code can be found in \url{https://github.com/coco11563/Traceable_Automatic_Feature_Transformation_via_Cascading_Actor-Critic_Agents}}

\end{abstract}
\vspace{-0.3cm}
\section{Introduction}
\vspace{-0.1cm}
Many applications and industrial sectors need to build ML systems. 
In practice, one of the essential steps in building an ML system is data preprocessing, transformation, and refinery. 
This is because, when data space is imperfect and low-quality, it is hard to develop an effective ML system, regardless of model fanciness. 
Fundamentally, this step can be generalized as a task of feature transformation, that is, transforming an original feature set into an optimized feature set that enhances the performances of a downstream ML model. 
Solving the feature transformation task can develop a more discriminative feature space, reconstruct contrastive pattern representations, improve traceability and explainability, and enhance downstream predictive performances. 

\begin{figure}[!htbp]
\vspace{-0.2cm}
    \centering
    \includegraphics[width=\linewidth]{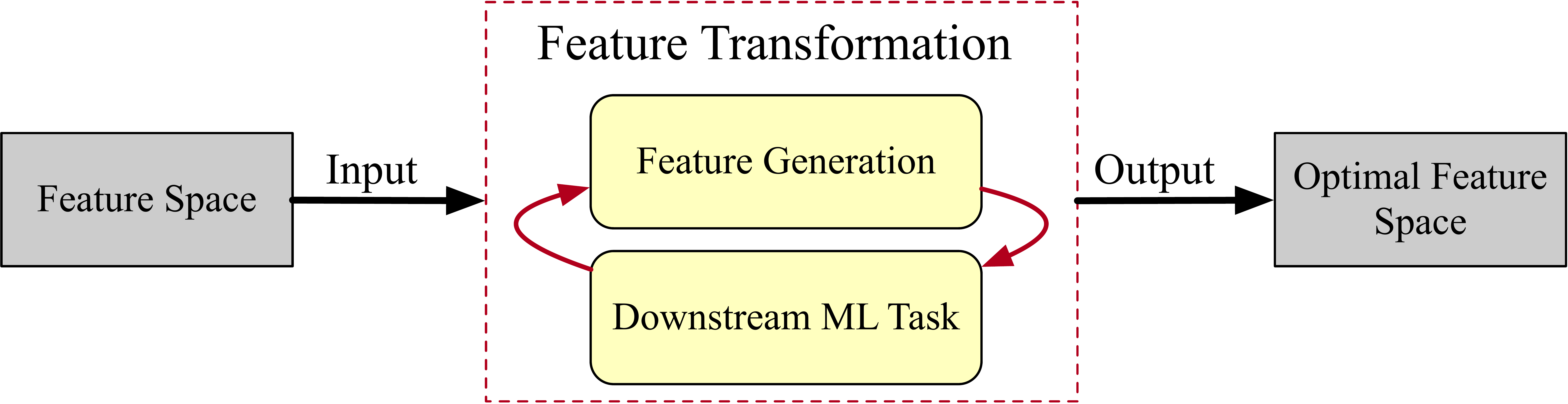}
    \vspace{-0.55cm}
    \caption{Given the input feature space, the automatic feature transformation task aims to output an optimal feature space via iteration between feature generation and feature selection.}
    \label{overall_task}
    \vspace{-0.45cm}
\end{figure}

One typical strategy of existing systems (Figure~\ref{overall_task}) is to iterate feature generation and selection to transform and refine the original feature space. 
More recent studies focused on automating such transformation tasks by machine intelligence, such as reinforcement learning~\cite{zhang2019automatic,chen2019neural,wang2022group}, to decide features and operations for crossing, generation, and subsetting.

There are three challenges in existing systems:
\emph{Issue 1: Overcoming the curse of dimensionality.}
Previous studies, such as~\cite{wang2022group},  used Deep Q-networks (DQN) to select candidate features and operations for feature space reconstruction. 
DQN regards all the candidate features to generate as action space and has to estimate the Q-value of each candidate.  
However, when an original feature set is big, the action space for agents will exponentially grow,  and the learning process is computationally costly.
The key question to answer is: how can we propose a better autonomous framework for both low-dimensional and high-dimensional feature space?
\emph{Issue 2: accurate state perception of uncertain feature space.} 
An intuitive way is to use the features in a feature set as descriptive statistics of the state of a feature space~\cite{khurana2018feature,chen2019neural}.
But, notice that a feature set varies over time during the transformation process; a feature set could include imperfect and redundant features; the fine-grained state of a feature space will vary even when a feature or operation is selected for the generation. As a result, it is challenging to perceive the state and structure of feature space for learning unbiased policies.
The key question to answer is: how can we learn effective and accurate state representation to improve the comprehension of reinforced agents?
\emph{Issue 3: generation with the trade-off between efficiency and global crossing among features.}
If we adopt a path-like generation strategy based on a  single feature's locally generated candidates~\cite{chen2019neural},  we will miss the globally optimized feature space.
If we adopt a globally crossing strategy, we sacrifice efficiency. 
How can we efficiently generate features while considering the structure information of a feature space?

\noindent\textbf{Our Insights: a group-wise cascading actor-critic perspective.}
We formulate the task of feature transformation as an iterative, nested process of feature generation and selection, where generation is to add new features, and selection is to reduce unnecessary features. 
We show that cascading actor-critic agents can learn more robust and accurate policies even under high-dimensional feature space with large action space. 
We highlight that group-wise feature crossing based generation can generate features efficiently while maintaining more global feature-feature crossing interactions. 
We found that statistics, autoencoder, and graph are three effective perspectives to perceive the situational state of varying and uncertain feature spaces. 
We demonstrate that our method can strategically unify the above three insights into a technical learning framework. 


\noindent\textbf{Summary of Proposed Approach.} Inspired by these findings, we propose a novel t\textbf{R}aceable \textbf{A}utomatic \textbf{F}eature \textbf{T}ransformation (RAFT) framework. The framework has three goals: 1)  \textbf{autonomous feature generation via cascading actor-critic agents.}  We use three actor-critic agents to select candidate features and operations to conduct feature crossing and generate new features. The actor component learns the probability distribution of actions through policy gradient update and directly outputs the selected action by sampling. This strategy can avoid estimating the Q-value of each action and, thus, accelerate large feature space transformation. The critic through uses gradient signals from temporal-difference errors at each iteration to evaluate and enhance selection policies. The collaboration between actor and critic components can converge quickly model convergence and learn robust generation policies.
2) \textbf{accurate state representation.} An effective state representation should describe not only the key dimensions of feature space but also model interconnected feature-feature correlations. 
To achieve this goal, we propose three advanced state representation methods from the statistic, encoding-decoding, and graph embedding perspectives.
3) \textbf{feature clustering and fast group-wise crossing.} We adopt a group-wise feature group-group crossing to generate a large number of features. We propose to cross feature groups with less information overlap in order to generate more informative dimensions. To maximize feature dissimilarity across groups, we develop a novel feature-feature distance metric for hierarchical feature grouping. 
Finally, the experimental results on 17 datasets with 3 application scenarios illustrate 24.7\% improvements in F1 scores and better robustness in high-dimensional data compared with SOTAs. 

\section{Problem Formulation}

\noindent\textbf{Feature Set and Target:}
We aim to reconstruct the feature space of such datasets $\mathcal{D}<\mathcal{F},y>$. 
Here, $\mathcal{F}$ is a feature set in which each column denotes a feature, and each row denotes a data sample;
$y$ is the target label set corresponding to samples.
To efficiently produce new features, we divide the feature set $\mathcal{F}$ into different feature groups via clustering, denoted by $\mathcal{C}$.
Each feature group is a subset of  $\mathcal{F}$.

\noindent\textbf{Operation Set:}
We perform a mathematical operation on existing features in order to generate new ones.
The collection of all operations is an operation set, denoted by $\mathcal{O}$.
There are two types of operations: unary and binary.
The unary operations include ``square'', ``exp'', ``log'', and etc.
The binary operations are ``plus'', ``multiply'', ``divide'', and etc.

\noindent\textbf{Cascading Agent.}
We develop a new cascading agent structure for feature generation. 
This structure is made up of  three agents: two feature group agents and one operation agent. 
They share state information and sequentially select feature groups and operations.

\noindent\textbf{Problem Statement:}
Our work aims to reconstruct an optimal and traceable feature space to improve downstream ML tasks through mathematically transforming original features.
 Formally, given a dataset $D<\mathcal{F},y>$, an operator set $\mathcal{O}$, and a downstream ML task $A$ (e.g., classification, regression, outlier detection), our purpose is to obain an optimal feature space $\mathcal{F}^{*}$ that maximizes the performance indicator $P$ of the task $A$ through iterative feature transformation. 
The optimization goal can be defined as follows:
    \vspace{-0.1cm}
\begin{equation}
\label{objective}
    \mathcal{F}^{*} = argmax_{\mathcal{\hat{F}}}( P_A(\mathcal{\hat{F}},y)),
    \vspace{-0.2cm}
\end{equation}
where $\mathcal{\hat{F}}$ can be viewed as a subset of a combination of  the original feature set $\mathcal{F}$ and the generated new features $\mathcal{F}^g$, and $\mathcal{F}^g$ is produced by applying the operations $\mathcal{O}$ to  the original feature set $\mathcal{F}$.

\vspace{-0.3cm}
\section{Methodology}
\begin{figure*}[!t]
    \centering
    \includegraphics[width=\linewidth]{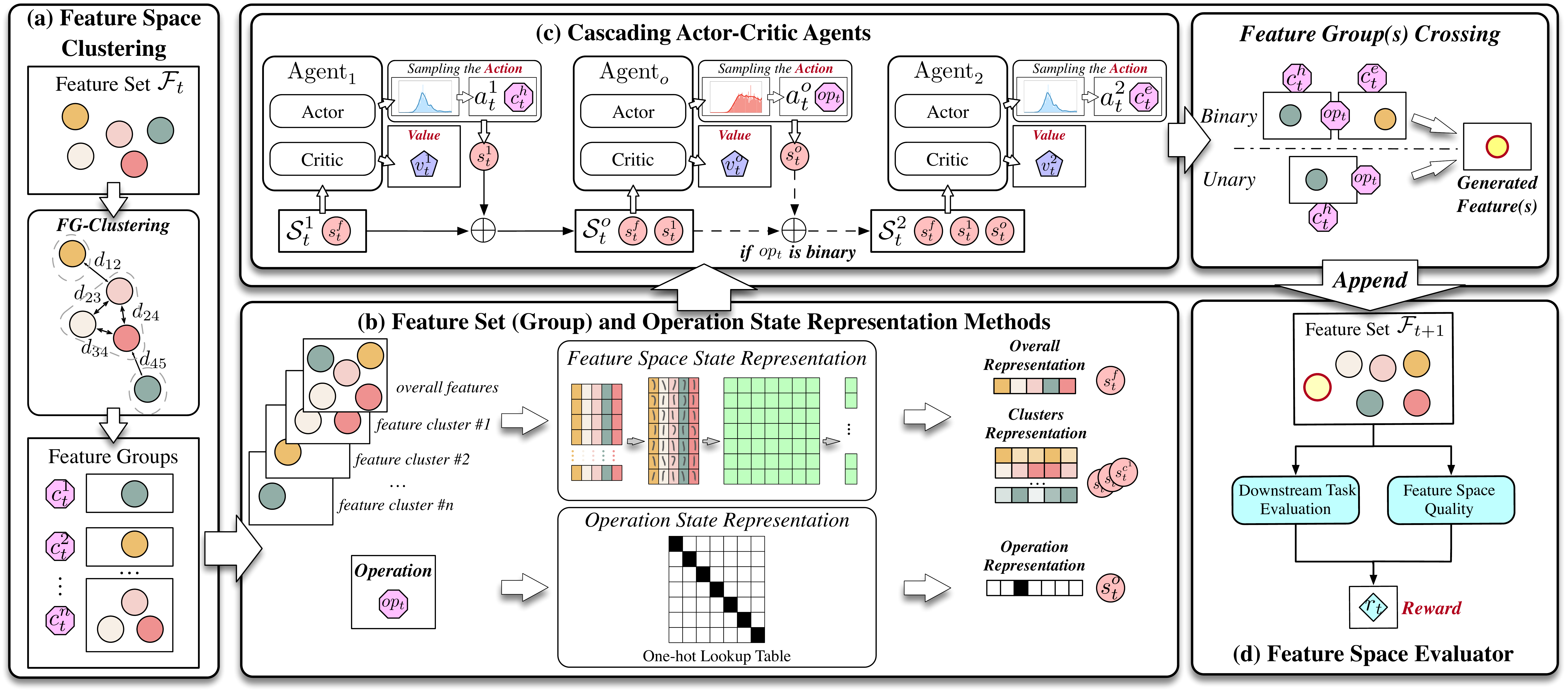}
    \vspace{-0.3cm}
    \captionsetup{justification=centering}
    \vspace{-0.33cm}
    \caption{An overview of the proposed framework RAFT in $t$-th iteration. (a) aims to cluster the input feature set. (b) aims to extract the state representations and help the cascading agents understand the current feature space. (c) aims to select the mathematical transformation. (d) aims to evaluate the generated feature set and obtain the overall reward.}
    \vspace{-0.6cm}
\label{model_overall}
\end{figure*}

We present the proposed framework RAFT as illustrated in Figure~\ref{model_overall}.
We demonstrate the technical details of each component in the following sections.
\vspace{-0.3cm}

\subsection{Feature Space Clustering}
To efficiently reconstruct feature space and provide strong reward signals to agents, we propose a feature space clustering component.
It divides the feature set into different feature groups, which builds a foundation for conducting group-wise feature generation.

\smallskip
\noindent\textbf{Features-Group Distance Function:}
We propose this function to measure the similarity between two clusters of features. Suppose we have two clusters $c_i$ and $c_j$, the formal definition of the features-group distance function is given by:
\begin{equation}
\begin{aligned}
  \small
    \label{dis}
    \mathcal{D}&(c_i, c_j)= \\
    &\frac{1}{|c_i|\cdot|c_j|}
    \sum_{f_i\in c_i}\sum_{f_j\in c_j}d(f_i,f_j)|I(f_i,y)-I(f_j,y)|,  
\end{aligned}
\vspace{-0.2cm}
\end{equation}
where $d(\cdot)$ is a generic pair-wise distance function (e.g., the euclidean distance, cosine similarity, etc) and $I(\cdot)$ is pairwise mutual information (PMI). The left part of Equation~\ref{dis} (i.e., $d(f_i,f_j)$) aims to quantify the numeric difference between $f_i$ and $f_j$. The right part  (i.e., $|I(f_i,y)-I(f_j,y)|$) aims to quantify the relevance differences between distinct features $f_i, f_j$ and the target $y$. 
This function seeks to aggregate features with similar information and the same contribution to differentiating the target label.
Because our assumption is that high (low) informative features are generated by crossing more distinct (similar) features.

\smallskip
\noindent\textbf{Feature-Group (FG) Clustering:}
Variable feature space sizes make it inappropriate to employ K-means or density-based clustering techniques during feature generation.
We propose an FG-Clustering algorithm inspired by agglomerative clustering.
Specifically, given a feature set $\mathcal{F}_t$ at the $t$-th step, 
we first initialize each feature column in $\mathcal{F}_t$ as a cluster at the beginning.
Then, 
we use the features-group distance function to calculate the distance between any two feature clusters. 
After that, we merge the two closest clusters to generate a new cluster and remove the former ones. 
We reiterate this process until the smallest distance between any two clusters breaks a certain threshold. 
Finally, we cluster $\mathcal{F}_t$ into different feature groups, defined as $C_t=\{c_i\}_{i=1}^{|C|}$.

\vspace{-0.3cm}
\subsection{State Representation for Feature Space and Operation}
To help cascading agents understand the current feature space for effective policy learning, we need to extract meaningful information from the space and use it as the state representation.
The assumption is that an effective state representation must not only capture the knowledge of feature space but also comprehend the correlations between features.
To achieve this goal, we introduce three state representation methods from different perspectives.  
To ease description, in the following parts, suppose given the feature set $\mathcal{F}\in \mathbb{R}^{M\times N}$, where $M$ is the number of total samples, and $N$ is the number of feature columns. 


\smallskip
\noindent\textbf{Statistic Information (si):} 
We utilize the statistic information (\textit{i.e.} count, standard deviation, minimum, maximum, first, second, and third quartile) of the feature space as the state representation.
Specifically, we first obtain the descriptive statistics matrix  of $\mathcal{F}$ column by column. Then, we calculate the descriptive statistics of the outcome matrix row by row to obtain the meta descriptive matrix that shape is $\mathbb{R}^{7\times 7}$. Finally, we obtain the state representation by flatting the descriptive matrix obtained from the former step. 
The state representation is defined as  $\mathcal{Z}_{si}(\mathcal{F})\in \mathbb{R}^{1\times 49}$.


\smallskip
\noindent\textbf{Autoencoder (ae):}
We propose an autoencoder-based state representation approach.
We believe an efficient state representation can reconstruct the original feature space.
Specifically, we first apply an autoencoder to transform each column of $\mathcal{F}$ into a latent matrix $Z \in \mathbb{R}^{k\times N}$, where $k$ is the dimension of the latent representation of each column.
Then, we apply another autoencoder to transform each row of $Z$ into another matrix $Z' \in \mathbb{R}^{k\times d}$, where $d$ is the dimension of the latent representation of each row.
After that, we want to use  $Z'$ to reconstruct the original feature space $\mathcal{F}$.
When the model converges, we flat  $Z'$ into one-dimensional vector and regard it as the state representation, denoted by  $\mathcal{Z}_{ae}(\mathcal{F}) \in \mathbb{R}^{1 \times kd}$.

\smallskip
\noindent\textbf{Graph Autoencoder (gae):}
In addition to reconstructing the feature space, we expect to preserve feature-feature correlations in the state representation.
Thus, we propose a graph autoencoder~\cite{https://doi.org/10.48550/arxiv.1611.07308} based state representation approach. 
Specifically, we first build a complete correlation graph $\mathcal{G}$ by calculating the similarity between each pair of feature columns.
The adjacency matrix of $\mathcal{G}$ is $\mathcal{A} \in \mathbb{R}^{N\times N}$, where a node is a feature column in $\mathcal{F}$ and an edge reflects the similarity between two nodes.
Then, we adopt a one-layer GCN~\cite{kipf2016semi} to aggregate feature knowledge of $\mathcal{F}$ based on $\mathcal{A}$ to produce an enhanced feature embedding $Z\in \mathbb{R}^{N\times k}$, where $k$ is the dimension of latent embedding. 
The calculation process is defined as follows:
$
Z = ReLU(\mathbf{D}^{-\frac{1}{2}}\mathcal{A}\mathbf{D}^{-\frac{1}{2}}\mathcal{F}^{\top}\mathbf{W}),
$
where $\mathbf{D}$ is the diagonal  degree matrix of $\mathcal{A}$, and $\mathbf{W}\in\mathbb{R}^{N\times k}$ is the weight matrix of the GCN.
Finally, we average $Z$ column-wisely to obtain the state representation, denoted by  $\mathcal{Z}_{gae}(\mathcal{F}) \in \mathbb{R}^{1\times k}$. 

For the selected mathematical operation, we use its one-hot vectors as the state representation, denoted by   $\mathcal{Z}_{o}(op)\in \mathbb{R}^{1\times |\mathcal{O}|}$. 

\vspace{-0.2cm}
\subsection{Feature Space Evaluator} 
We evaluate the quality of feature space and provide reward signals to reinforced agents to let them learn better feature transformation policies.
We assess the feature space from the following two feature utility perspectives:

\smallskip
\noindent\textbf{Downstream Task Evaluation:} 
We utilize the improvement of a downstream task (e.g., regression, classification, outlier detection) as one feature utility measurement. In detail, we use a downstream ML task with a task-specific indicator (e.g., 1-RAE, Precision, Recall, F1) to obtain the downstream task performance on the feature space.
The performance is denoted by  $P_A(\mathcal{F}, y)$. 

\smallskip
\noindent\textbf{Feature Space Quality:}
We also expect that the generated feature space should contain less redundant information and be more relevant to the target label.
Thus, we customize a feature space quality metric based on mutual information, which is defined as:
\begin{equation}
\small
    U(\mathcal{F}|y) = -\frac{1}{|\mathcal{F}|^2}\sum_{f_i, f_j \in \mathcal{F}} I(f_i, f_j) + \frac{1}{|\mathcal{F}|}\sum_{f\in \mathcal{F}}I(f,y),
\end{equation}
where $f_i, f_j, f$ are distinct features in $\mathcal{F}$, $I$ refers to the mutual information function,  and $|\mathcal{F}|$ is the size of the feature set $\mathcal{F}$.

\subsection{Cascading Agents} 
To intelligently select suitable features and operations for feature crossing, we decompose the selection process into three Markov Decision Processes (MDPs).
They cascade and sequentially select the first feature cluster, mathematical operation, and the second feature cluster. 
We develop a cascading actor-critic agent structure to make sure that all three agents collaborate with each other.
Figure~\ref{model_overall}(c) shows the model structure.
To ease the description, we adopt the $t$-th iteration as an example to illustrate the calculation process. 
Assuming that the feature set is  $\mathcal{F}_t$ and its feature clusters $\mathcal{C}_t$, we aim to obtain the next new feature space $\mathcal{F}_{t+1}$ by generating new features $g_t$.

\smallskip
\noindent\textbf{First Feature Cluster Agent:}
$\text{Agent}_1$  is to select the first candidate feature group. 
Its learning system includes the following:
\ul{\textit{State}:} the state  is the embedding vector of the current feature space $\mathcal{F}_t$, denoted by  $\mathcal{S}^1_t=s^f_t$, where  $s^f_t=\mathcal{Z}(\mathcal{F}_t)$. 
\ul{\textit{Action}:} the action is the first candidate feature group $c^h_t$ selected by
$\text{Agent}_1$ from $\mathcal{C}_t$, denoted by  $a^1_t=c^h_t$. 
\ul{\textit{Reward}:} the reward is the feature space quality score of the selected first feature group, dented by $r^1_t = U(c^h_t|y)$. 

\smallskip
\noindent\textbf{Operation Agent:} 
$\text{Agent}_o$ is to select a candidate mathematical operation. 
Its learning system includes: \ul{\textit{State}:} the state is the combination of the current feature space $\mathcal{F}_t$ and the selected first feature group $c^h_t$, denoted by  $\mathcal{S}^o_t=s^f_t\oplus s^{1}_t$, where $\oplus$ is the a row-wise concatenation and  $s^{1}_t = \mathcal{Z}(c^h_t)$. 
\ul{\textit{Action}:} the action is the candidate operation $op_t$  selected by $\text{Agent}_o$  from the operation set $\mathcal{O}$, denoted by $a^o_t=op_t$.
\ul{\textit{Reward}:} the reward is the integration of performance improvements of the downstream task and the quality score of the new generated feature space, denoted by  $r^o_t = U(\mathcal{F}_{t+1}|y) + P_A(\mathcal{F}_{t+1}, y) - P_A(\mathcal{F}_t, y)$. 

\smallskip
\noindent\textbf{Second Feature Cluster Agent:}
Agent$_2$ is to select the second candidate feature group.
\ul{\textit{State}:} the state is the combination of the embedding of the current feature space $s^f_t$, the first candidate feature group $s^{1}_t$, and the selected operation $s^o_t$, denoted by $\mathcal{S}^2_t=s^f_t\oplus s^{1}_t\oplus s^o_t$, where $s^o_t=\mathcal{Z}_o(op)$.
\ul{\textit{Action}:} the action is the second candidate feature group selected by Agent$_2$ from $\mathcal{C}_t$, denoted by $a^2_t=c^l_t$.
\ul{\textit{Reward}:} the reward is the quality score of the new generated feature space $\mathcal{F}_{t+1}$, denoted by $r^2_t = U(\mathcal{F}_{t+1}|y)$

\smallskip
\noindent\textbf{Feature Group(s) Crossing:}
After we have two candidate feature groups and one operation, we need to cross feature groups to create new features for refining feature space.
Based on the type of operation $op$, we propose two feature generation strategies to generate new features $g_t$.
\vspace{-0.3cm}
\begin{equation}
\vspace{-0.2cm}
    g_t = \begin{cases}
    op_t(c_t^1) : \text{if }op_t\text{ is unary}\\
    op_t(c_t^1, c_t^2) : \text{if }op_t\text{ is binary}
    \end{cases}.
\end{equation}
Specifically, if  $op$ is unary (\textit{e.g.,} square, sqrt), we conduct it on the first selected feature group; if $op$ is binary (\textit{e.g.,} plus, divide), we apply it to the two candidate feature groups.
Then, $g_t$ is added into the $\mathcal{F}_t$ to form the new feature set $\mathcal{F}_{t+1}$.
If the feature space size exceeds a maximization threshold, redundant features are eliminated using feature selection to control the feature space size.
We reiterate the feature transformation process until finding the optimal feature set $\mathcal{F}^*$ or achieving the maximum iteration number.

\vspace{-0.2cm}
\subsection{Actor-Critic Optimization Strategy} 
We adopt the same training strategy (actor-critic) to train the three agents in order to learn smart and ideal feature transformation policies.

\smallskip
\noindent The actor-critic paradigm consists of two components: 

\smallskip
\noindent\ul{\textit{Actor}}: The actor aims to learn the selection policy  $\pi(\cdot)$ based on the current state in order to select suitable candidate feature groups or operations.
 In the $t$-th iteration, with given state $\mathcal{S}_t$, the agent will pick an action $a_t$, defined as:
\begin{equation}
\label{actor}
    a_t \sim \pi_\theta(\mathcal{S}_{t}),
\end{equation}
where $\theta$ is the parameter of policy network $\pi$. The output of the $\pi_\theta(\cdot)$ is the probability of each candidate action. $\sim$ operation means the sampling operation.  

\smallskip
\noindent\ul{\textit{Critic}}:
The critic aims to estimate the potential reward of an input state, given by:
\begin{equation}
    v_t = V(\mathcal{S}_t),
\end{equation}
where $V(\cdot)$  is the state-value function and $v_t$ is the obtained value. 

\smallskip
\noindent We update the \textit{Actor} and \textit{Critic} in cascading agents after each iteration of feature transformation.  
Suppose at the $t$-th iteration, for one agent, we can obtain the memory as $\mathcal{M}_t = (a_t, \mathcal{S}_t, \mathcal{S}_{t+1}, r_t)$. 
Then, the formal definition of policy gradient is given by:
\begin{equation}
    \nabla J(\theta)_t = \nabla_\theta log\pi_\theta(a_t|\mathcal{S}_t)(Q(\mathcal{S}_t, a_t) - V(\mathcal{S}_t)),
    \label{pg}
\end{equation}
where $(Q(\mathcal{S}_t, a_t) - V(\mathcal{S}_t))$ is the advantage function ($\delta$).  $\pi (a_t|\mathcal{S}_t)$ denote the probability of selected action $a_t$. $Q(\mathcal{S}_t, a_t)$ can be estimated by the state-value function (i.e., \textit{Critic}) and the reward of the current step, which is defined as:
\begin{equation}
Q(\mathcal{S}_t, a_t)\approx r_t + \gamma V(\mathcal{S}_{t+1}).
\end{equation}
where $\gamma \in [0, 1]$ is the discounted factor.
During the training phase, suppose the RAFT has explored the feature transformation graph $n$ steps and collected the memories. Then, we optimize the parameter of \textit{Critic} to provide a more precise state-value estimation by minimizing this: 
\begin{equation}
    \mathcal{L}_c = \frac{1}{n}\sum_{i=1}^n(r_i + \gamma V(\mathcal{S}_{i+1}) - V(\mathcal{S}_i))^2.
\end{equation}
After that, we optimize the policy of  \textit{Actor} based on Equation~\ref{pg}:
\begin{equation}
    \mathcal{L}_a = \frac{1}{n}\sum_{i=1}^n ( log\pi_\theta (a_i|\mathcal{S}_i) * \delta + \beta H(\pi_\theta (\mathcal{S}_i))),
\end{equation}
where $\delta$ is the advantage function. $H(\cdot)$ is an entropy regularization term that aims to increase the randomness in exploration. We use $\beta$ to control the strength of the $H$.  
The overall loss function for each agent is:
\begin{equation}
    \mathcal{L} = \mathcal{L}_c + \mathcal{L}_a.
\end{equation}
After agents converge, we expect to discover the optimal policy $\pi^*$ that can choose the most appropriate action (\textit{i.e.} feature group or operation).

\section{Experiment}
\begin{table*}[!h]
\centering
\caption{Overall Performance. The best results are highlighted in \textbf{bold}. The second-best results are highlighted in \underline{underline}. We annotate the performance improvement of RAFT compared with the original feature space.}
\vspace{-3mm}
\label{table_overall_perf}
\setlength{\tabcolsep}{2.0mm}{\resizebox{\linewidth}{!}{
\begin{tabular}{cccccccccccccccccc}
\toprule
Name & Source   & Task & Samples & Features & Data Std.& RFG  & {ERG} & LDA & AFT   & NFS   & TTG  & GRFG & RAFT \\  \midrule
PimaIndian      & UCIrvine & C   & 768     & 8 & High & 0.693 & 0.703 & 0.676  & 0.736 & 0.762 & 0.747 & \ul{0.776} & \textbf{0.789}$^{+6.9\%}$\\  
SVMGuide3  & LibSVM & C & 1243 & 21 & Mid  & 0.703 & 0.747 & 0.683 & 0.829 & 0.831 & 0.766 & \ul{0.850} & \textbf{0.858}$^{+5.0\%}$\\  
Amazon Employee    & Kaggle   & C   & 32769   & 9 & High & 0.744 & 0.740 & 0.920  & 0.943 & 0.935 & 0.806 & \ul{0.946} & \textbf{0.946}$^{+1.7\%}$\\  
German Credit      & UCIrvine & C   & 1001    & 24  & High     & 0.695 & 0.661 & 0.627  & 0.751 & 0.765 & 0.731 & \ul{0.772}& $\textbf{0.774}^{+4.7\%}$\\  
Wine Quality Red   & UCIrvine & C   & 999     & 12  & High    & 0.599 & 0.611 & 0.600  & 0.658 & 0.666 & 0.647 & \ul{0.686}& \textbf{0.697}$^{+3.6\%}$\\  
Wine Quality White & UCIrvine & C   & 4900    & 12   & High    & 0.552 & 0.587 & 0.571  & 0.673 & 0.679 & 0.638 & \ul{0.685}& \textbf{0.693}$^{+2.2\%}$\\   \midrule
Openml\_618        & OpenML   & R   & 1000    & 50  & Low   & 0.415 & 0.427  & 0.372 & 0.665 & 0.640 & 0.587 & \ul{0.672} & \textbf{0.803}$^{+20.7\%}$\\  
Openml\_589        & OpenML   & R   & 1000    & 25  & Low & 0.638 & 0.560 & 0.331  & 0.672 & 0.711 & 0.682 & \ul{0.753} & \textbf{0.782}$^{+16.1\%}$\\  
Openml\_616        & OpenML   & R   & 500     & 50 & Low   & 0.448 & 0.372 & 0.385  & 0.585 & 0.593 & 0.559 & \ul{0.603} & \textbf{0.717}$^{+21.9\%}$\\  
Openml\_607        & OpenML   & R   & 1000    & 50  & Low    & 0.579 & 0.406 & 0.376  & 0.658 & 0.675 & 0.639 & \ul{0.680} & \textbf{0.756}$^{+14.7\%}$\\  
Openml\_620        & OpenML   & R   & 1000    & 25  & Low   & 0.575 & 0.584 & 0.425 & 0.663 & 0.698 & 0.656 & \ul{0.714} & \textbf{0.720}$^{+10.5\%}$\\  
Openml\_637        & OpenML   & R   & 500     & 50   & Low     & 0.561 & 0.497 & 0.494  & 0.564 & 0.581 & 0.575 & \ul{0.589} & \textbf{0.644}$^{+15.2\%}$\\  
Openml\_586       & OpenML   & R   & 1000    & 25  & Low   & 0.595 & 0.546 & 0.472  & 0.687 & 0.748 & 0.704 & \ul{0.783} & \textbf{0.802}$^{+16.7\%}$\\  \midrule
WBC & UCIrvine   & D   & 278    & 30 & Low & 0.753 & 0.766 & 0.736  & 0.743 & 0.755 & 0.752 & \ul{0.785} & \textbf{0.979}$^{+31.4\%}$\\  
Mammography        & OpenML   & D   & 11183    & 6 & Low      & 0.731 & 0.728 & 0.668  & 0.714 & 0.728 & 0.734 & \ul{0.751} & $\textbf{0.832}^{+9.6\%}$\\  
Thyroid        & UCIrvine   & D   & 3772    & 6  & Mid      & 0.813 & 0.790 & 0.778  & 0.797 & 0.722 & 0.720 & \ul{0.954} & \textbf{0.998}$^{+17.8\%}$\\  
SMTP        & UCIrvine   & D   & 95156    & 3 & Mid       & 0.885 & 0.836 & 0.765  & 0.881 & 0.816 & 0.895 & \ul{0.943}& \textbf{0.949}$^{+16.2\%}$\\  \bottomrule
\end{tabular}}}
\vspace{-5mm}
\end{table*}
\begin{figure*}[!h]
\centering
\subfigure[PimaIndian]{
\includegraphics[width=4.0cm]{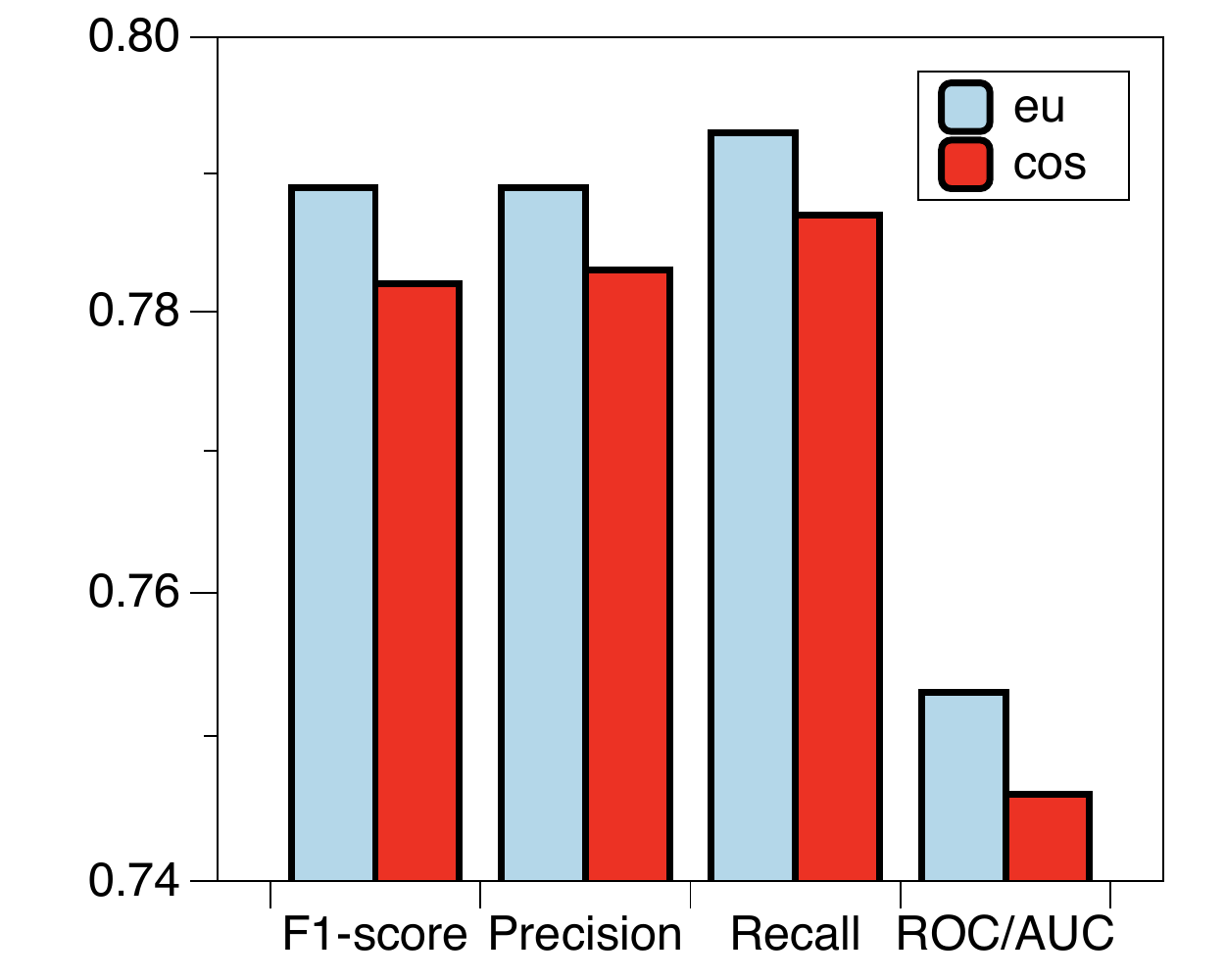}
}
\hspace{-3mm}
\subfigure[Wine Quality Red]{
\includegraphics[width=4.0cm]{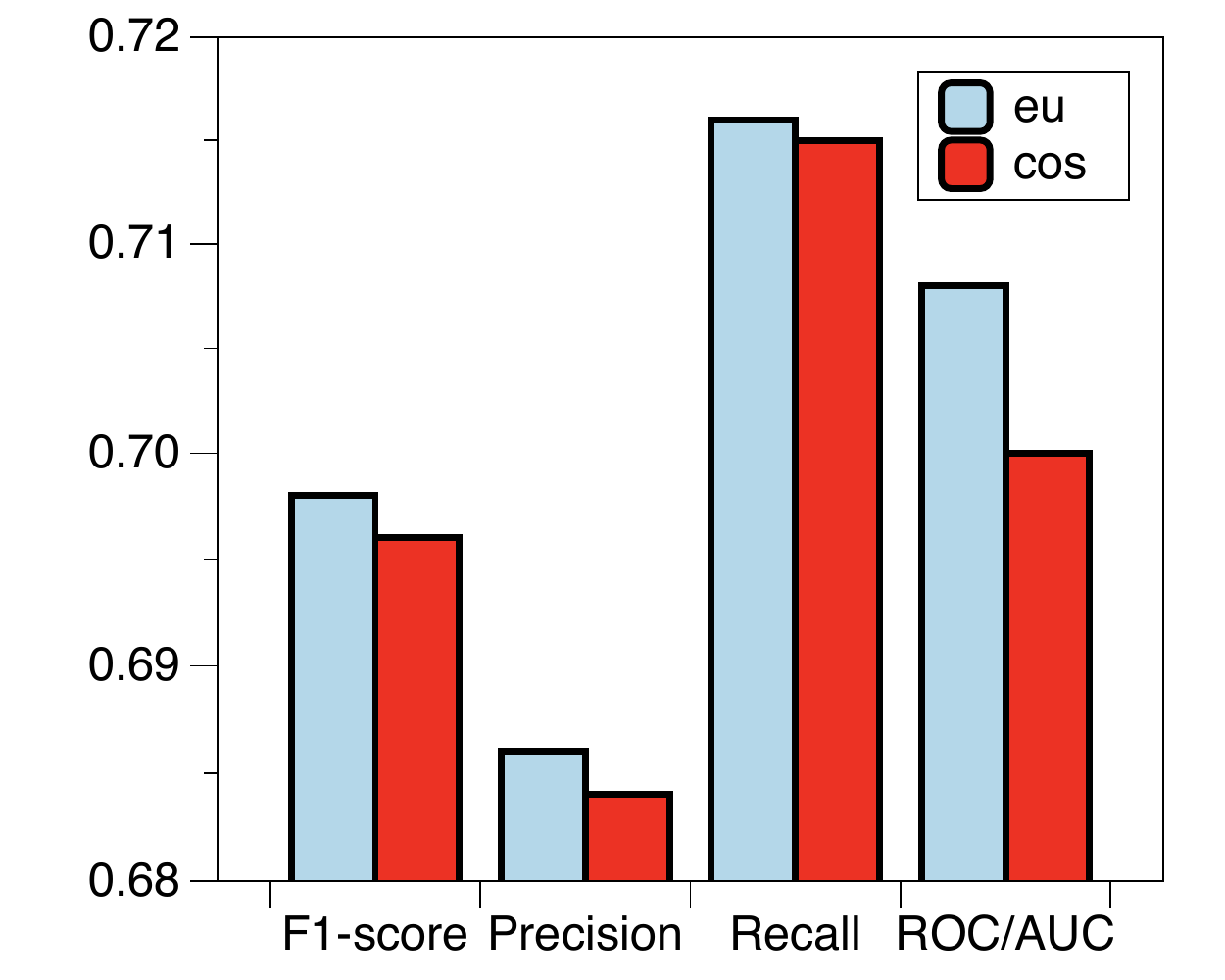}
}
\hspace{-3mm}
\subfigure[SVMGuide3]{ 
\includegraphics[width=4.0cm]{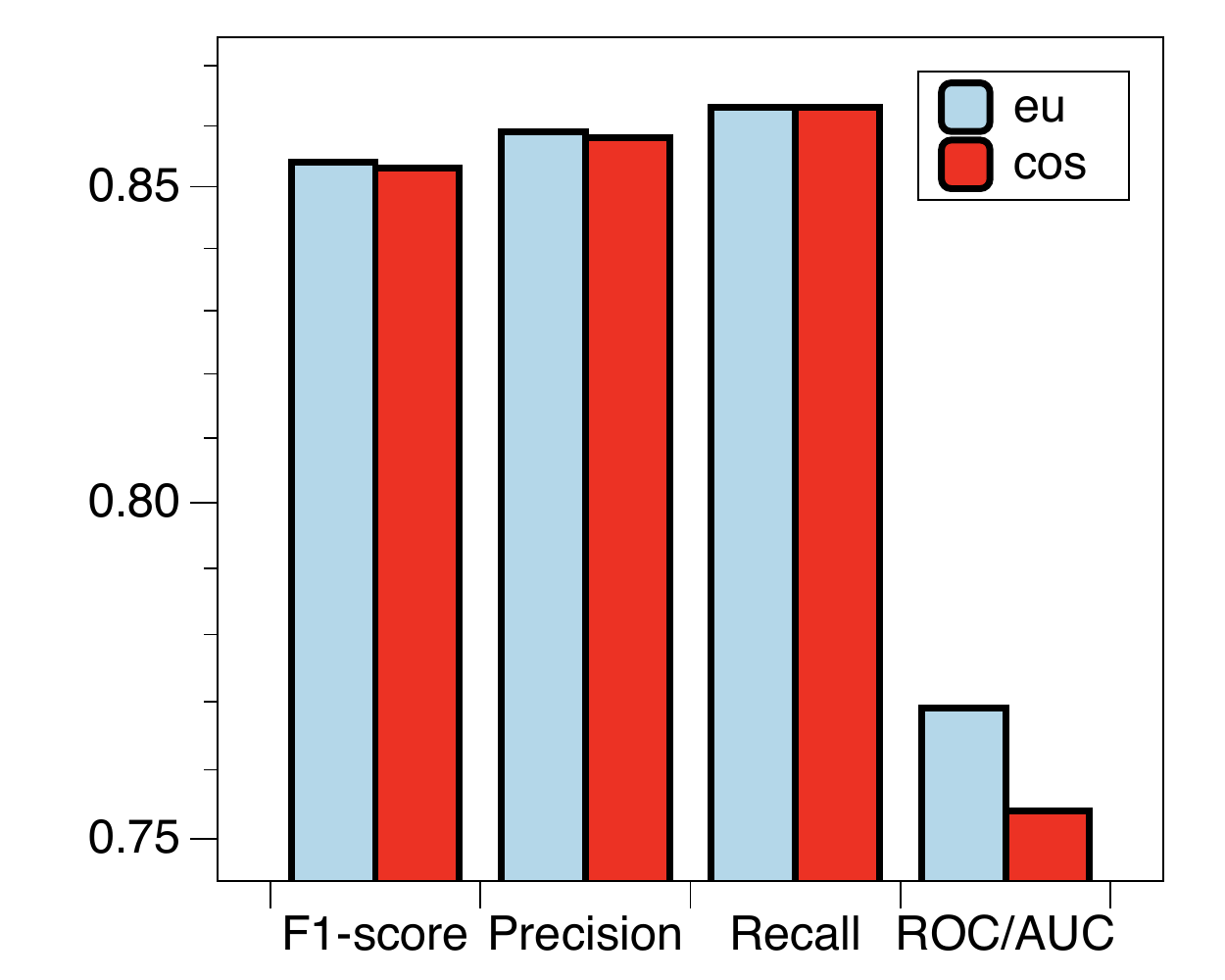}
}
\hspace{-3mm}
\subfigure[Thyroid]{ 
\includegraphics[width=4.0cm]{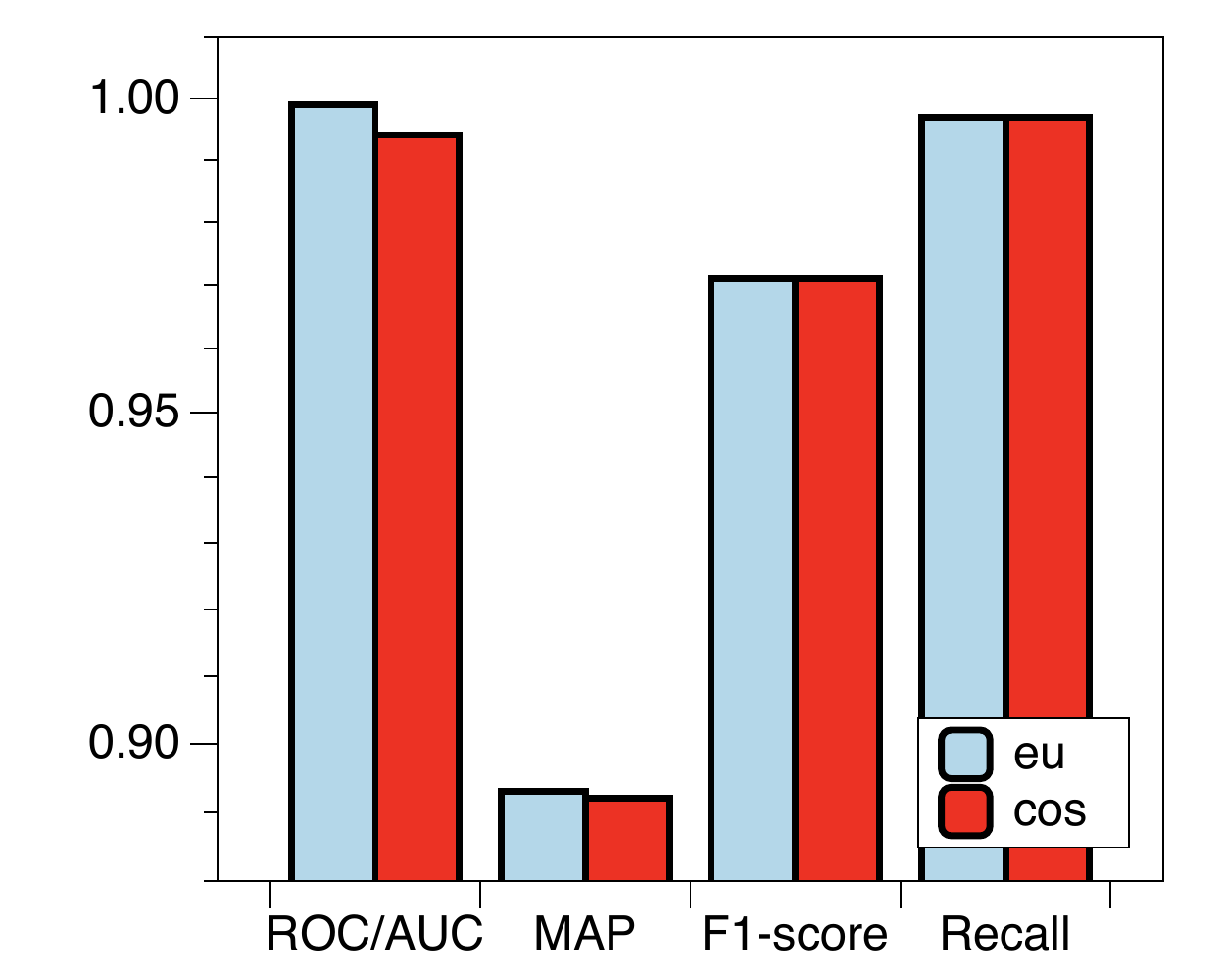}
}
\vspace{-2.5mm}

\subfigure[Openml\_586]{ 
\includegraphics[width=4.0cm]{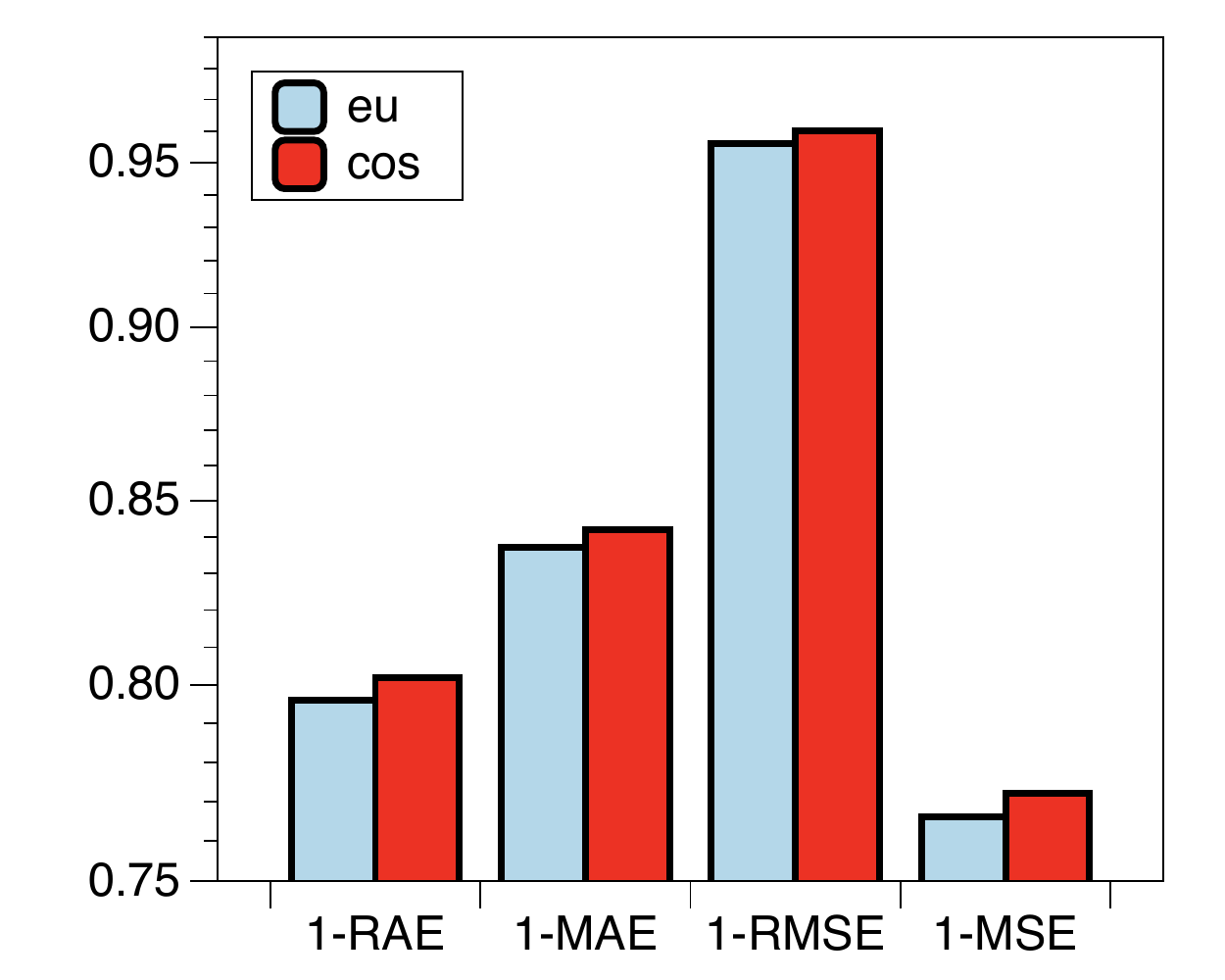}
}
\hspace{-3mm}
\subfigure[Openml\_618]{ 
\includegraphics[width=4.0cm]{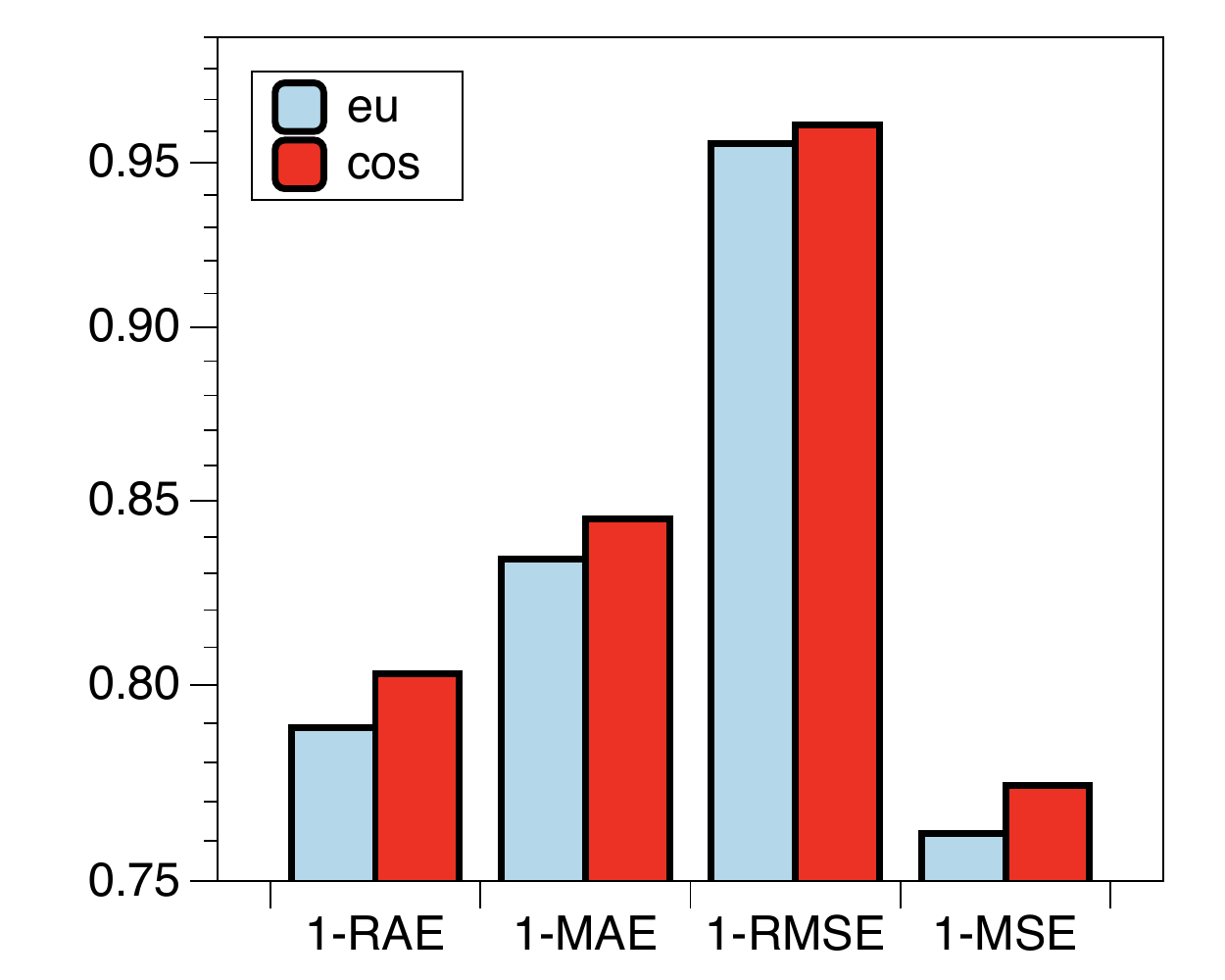}
}
\hspace{-3mm}
\subfigure[Mammography]{ 
\includegraphics[width=4.0cm]{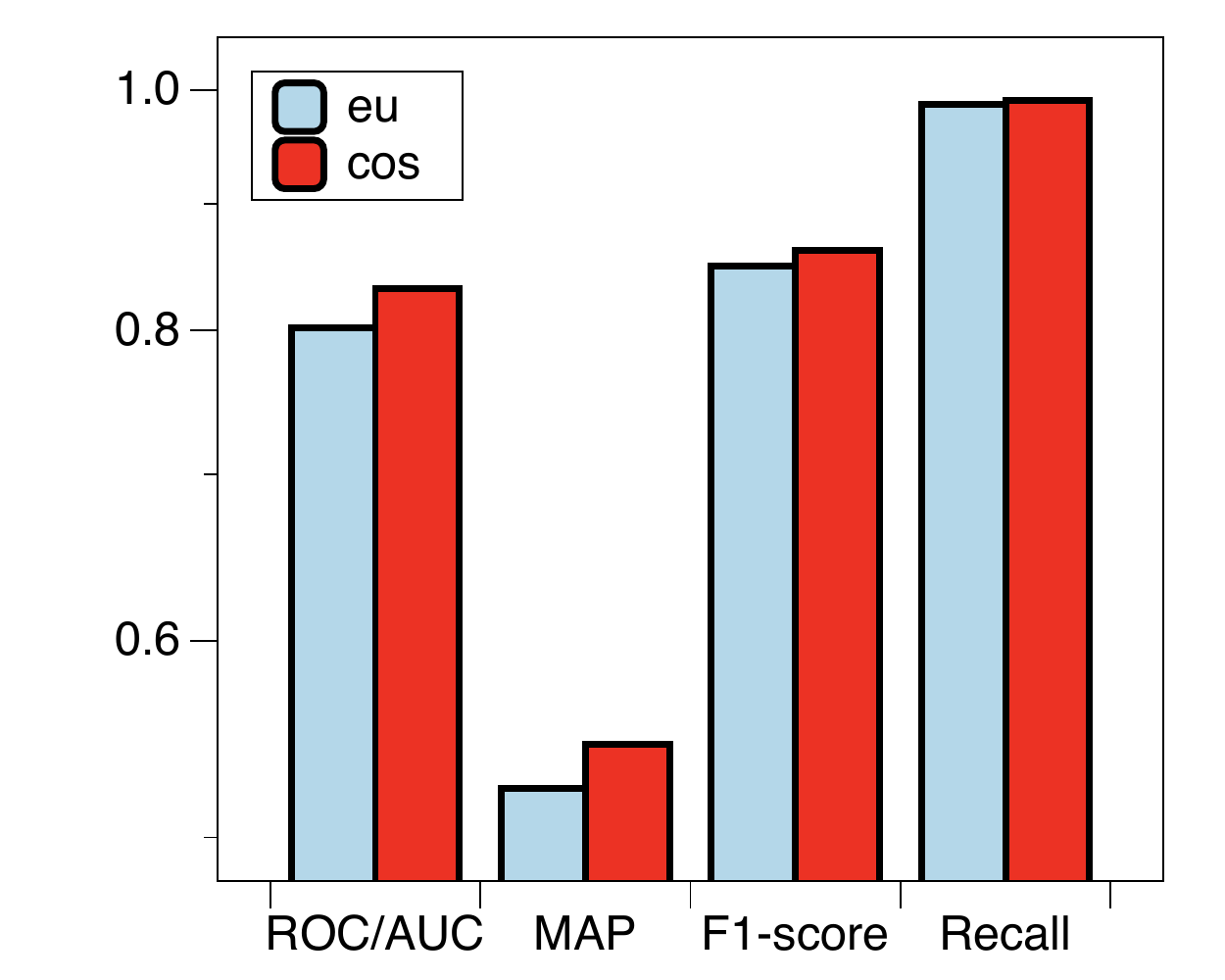}
}
\hspace{-3mm}
\subfigure[WBC]{ 
\includegraphics[width=4.0cm]{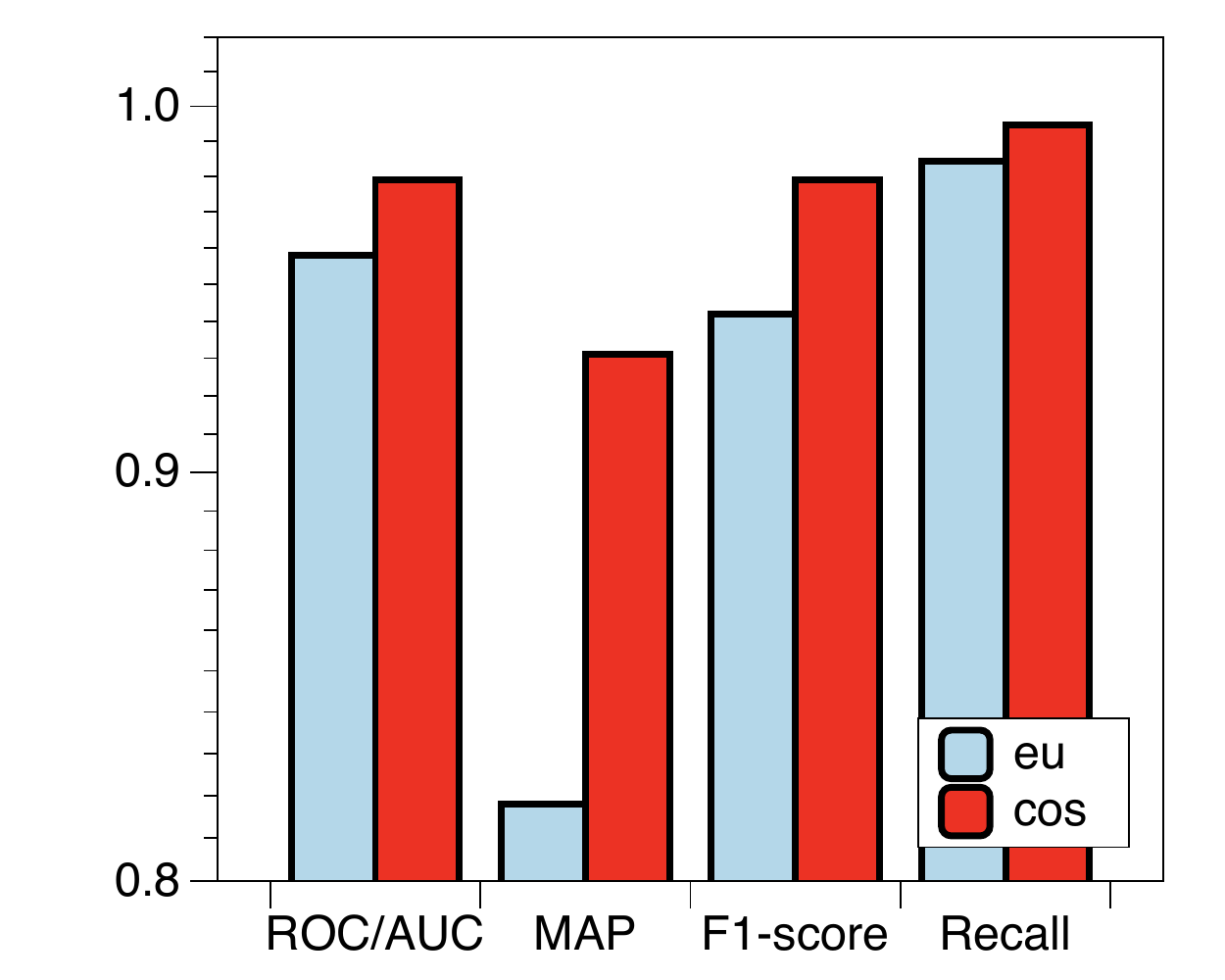}
}
\vspace{-0.35cm}
\caption{Comparison of different  distance functions using in FG-Clustering. }
\label{dis_study}
\vspace{-0.8cm}
\end{figure*}
\subsection{Data Description}
\vspace{-0.1cm}
We used 17 publicly available datasets from UCI~\cite{uci}, 
 LibSVM~\cite{libsvm},
 Kaggle~\cite{kaggle}, and OpenML~\cite{openml} to conduct experiments.
The 17 datasets involve 6 classification tasks, 7 regression tasks, and 4 outlier detection tasks.
Table \ref{table_overall_perf} shows the statistic information of these datasets.
We also categorized these datasets into \textit{High} (higher than 5), \textit{Mid} (between 0.01 to 5), and \textit{Low} (between 0 to 0.01)
based on the standard deviation of the feature set.
The dataset has a larger standard deviation, indicating that its value range is also larger, and vice versa.

\subsection{Evaluation Metrics}
\vspace{-0.3cm}
We used  F1-score, Precision,  Recall, and  ROC/AUC to evaluate  classification tasks. 
We used   1-Relative Absolute Error (1-RAE)~\cite{wang2022group},  1-Mean Average Error (1-MAE),  1-Mean Square Error (1-MSE), and  1-Root Mean Square Error (1-RMSE) to evaluate regression tasks. 
We adopted  ROC/AUC,  Mean Average Precision (MAP), F1-score, and  Recall to assess outlier detection tasks. 

\vspace{-0.3cm}
\subsection{Baseline Algorithms}
\label{baseline}
We compared our work RAFT with seven widely-used feature engineering methods:
(1) \textbf{RFG} randomly selects candidate features and operations for generating new features without any policy learning; 
(2) \textbf{ERG} is a expansion-reduction method, which applies operations to all features to expand the feature space, then  selects critical features as a new feature space. 
(3) \textbf{LDA}~\cite{blei2003latent} extracts latent features from the feature set via matrix factorization.
(4) \textbf{AFT}~\cite{horn2019autofeat}  is an enhanced ERG implementation that iteratively explores feature space and adopts multi-step feature selection to reduce redundant features.
(5) \textbf{NFS}~\cite{chen2019neural} mimics feature transformation path for each feature and optimizes the entire transformation process based on reinforcement learning.
(6) \textbf{TTG}~\cite{khurana2018feature} records the feature transformation process using a transformation graph, then uses reinforcement learning to explore the graph to determine the best feature set.
(7) \textbf{GRFG}~\cite{wang2022group} is an automatic feature generation method, which is optimized through DQN. 
\begin{figure*}[!h]
\centering
\subfigure[PimaIndian]{
\includegraphics[width=5.2cm]{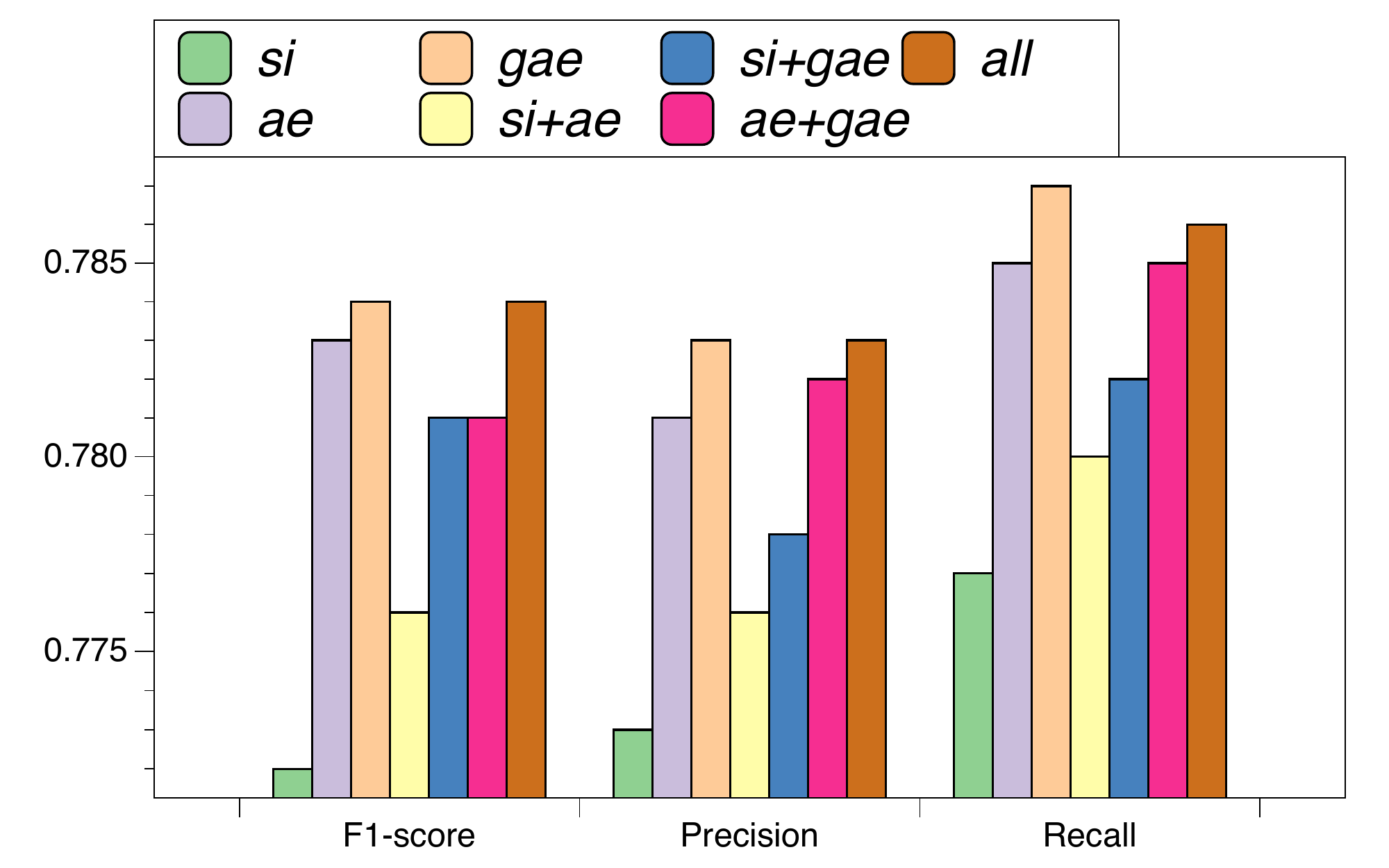}
}
\hspace{-3mm}
\subfigure[OpenML\_586]{ 
\includegraphics[width=5.2cm]{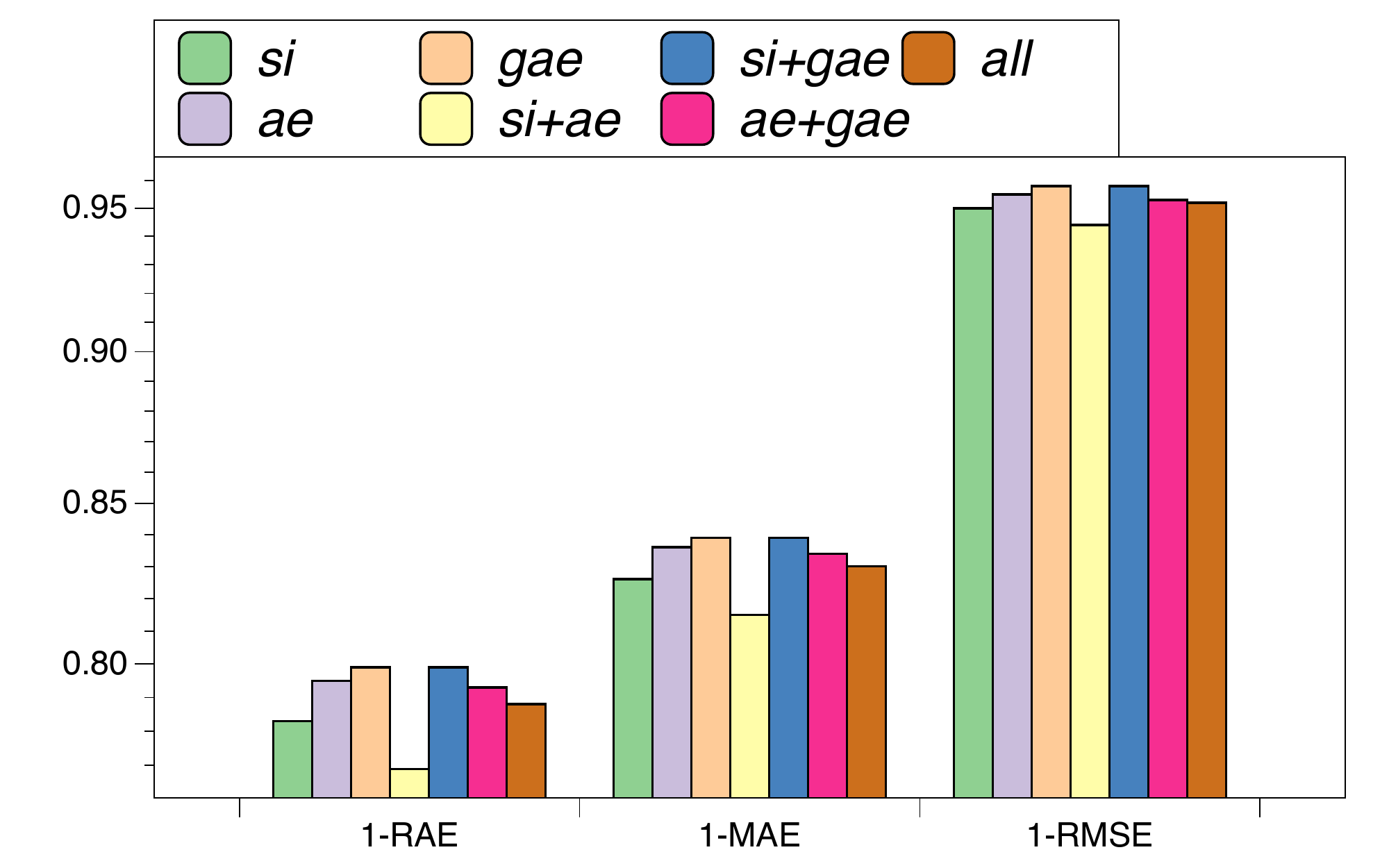}
}
\hspace{-3mm}
\vspace{-3mm}
\subfigure[WBC]{
\includegraphics[width=5.2cm]{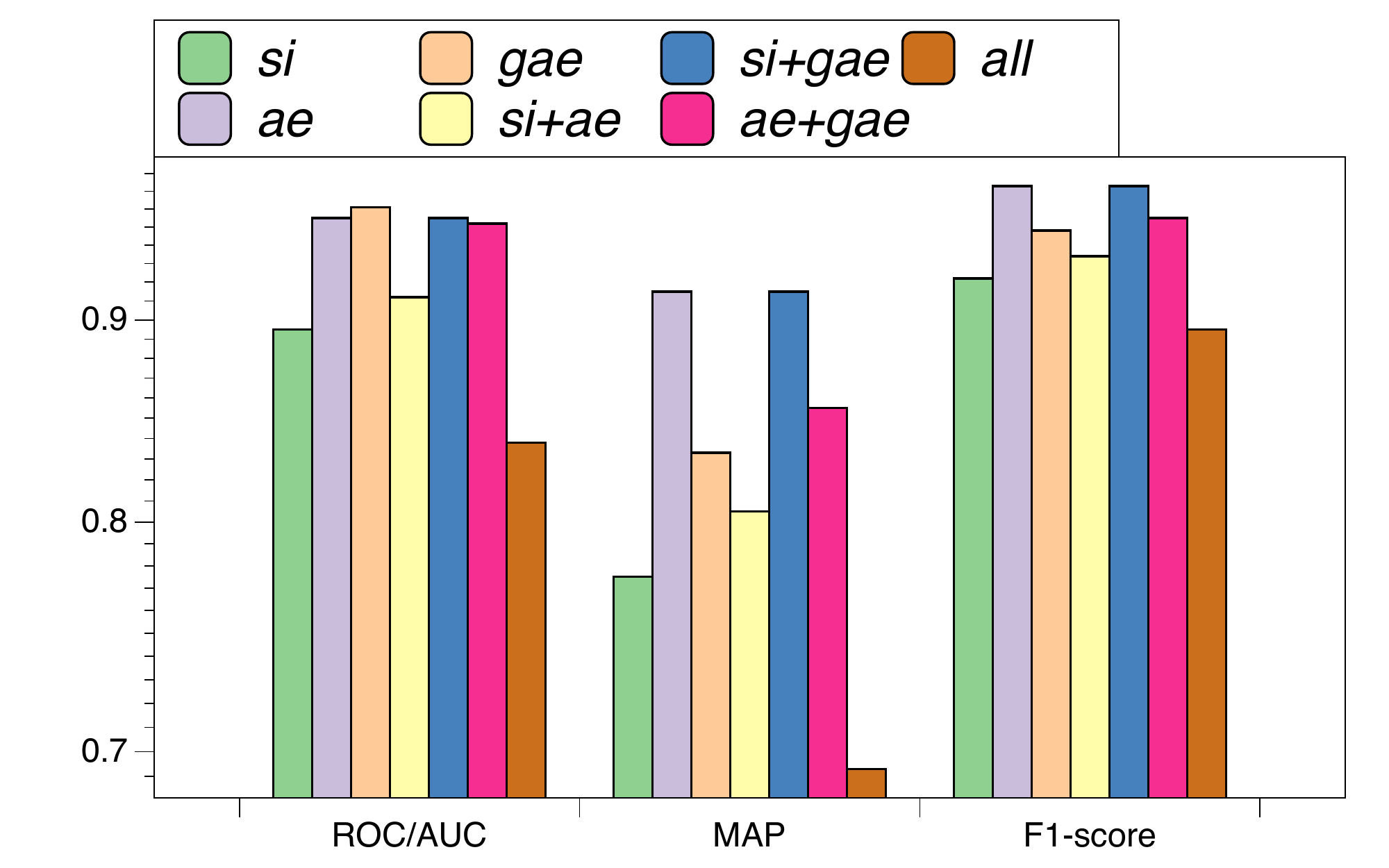}
}
\hspace{-3mm}
\subfigure[Wine Quality Red]{ 
\includegraphics[width=5.2cm]{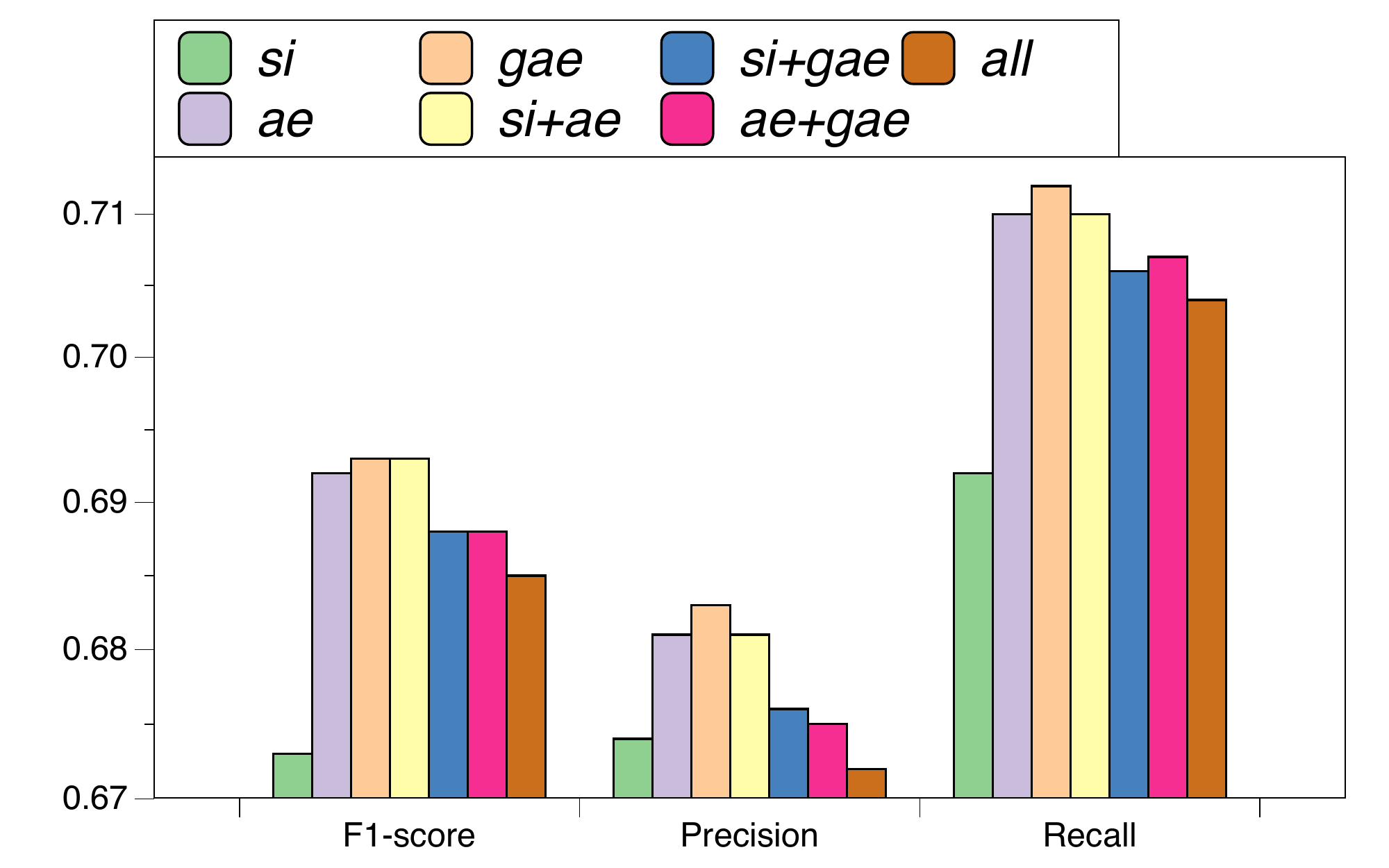}
}
\hspace{-3mm}
\subfigure[OpenML\_618]{ 
\includegraphics[width=5.2cm]{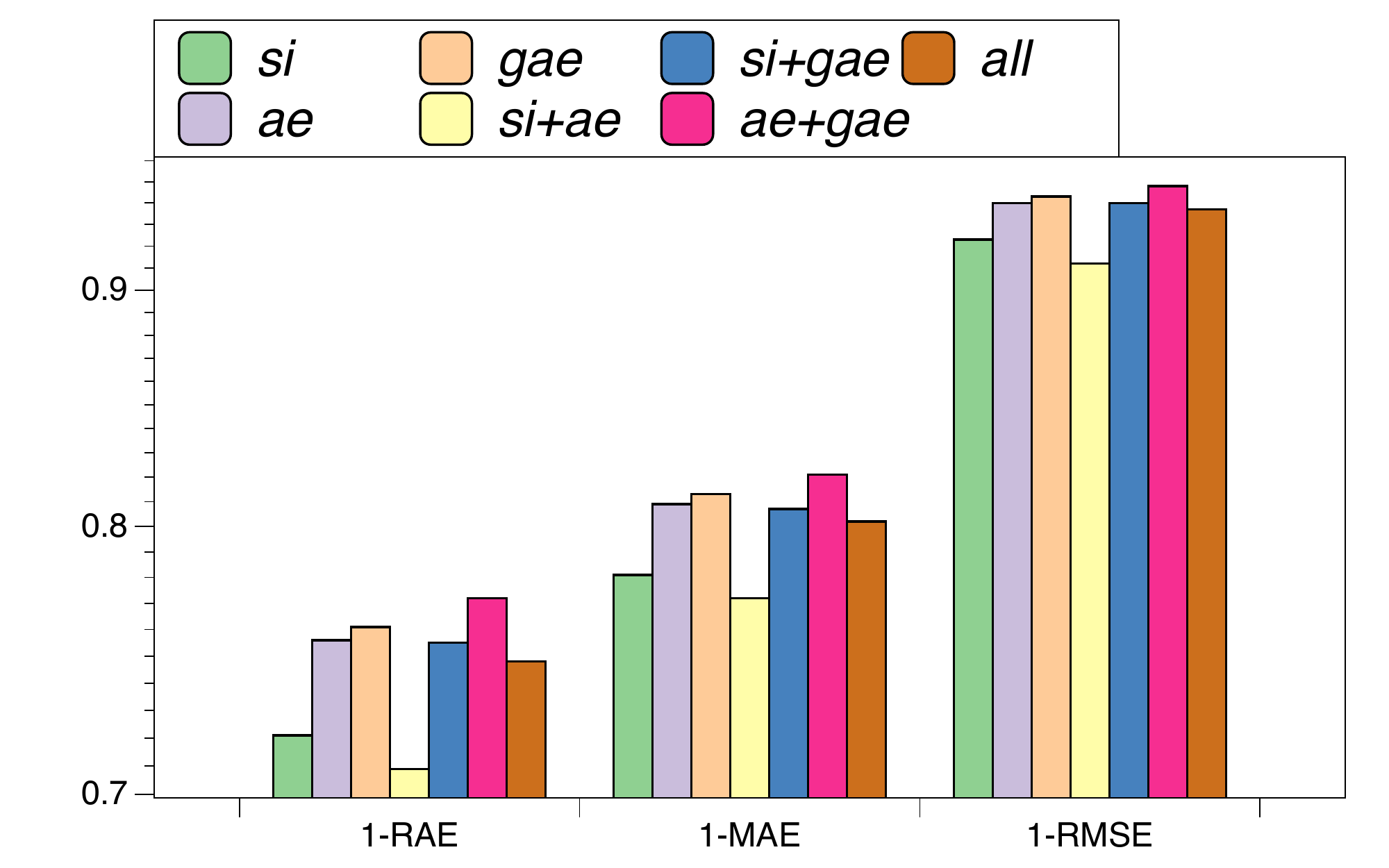}
}
\hspace{-3mm}
\subfigure[Thyroid]{ 
\includegraphics[width=5.2cm]{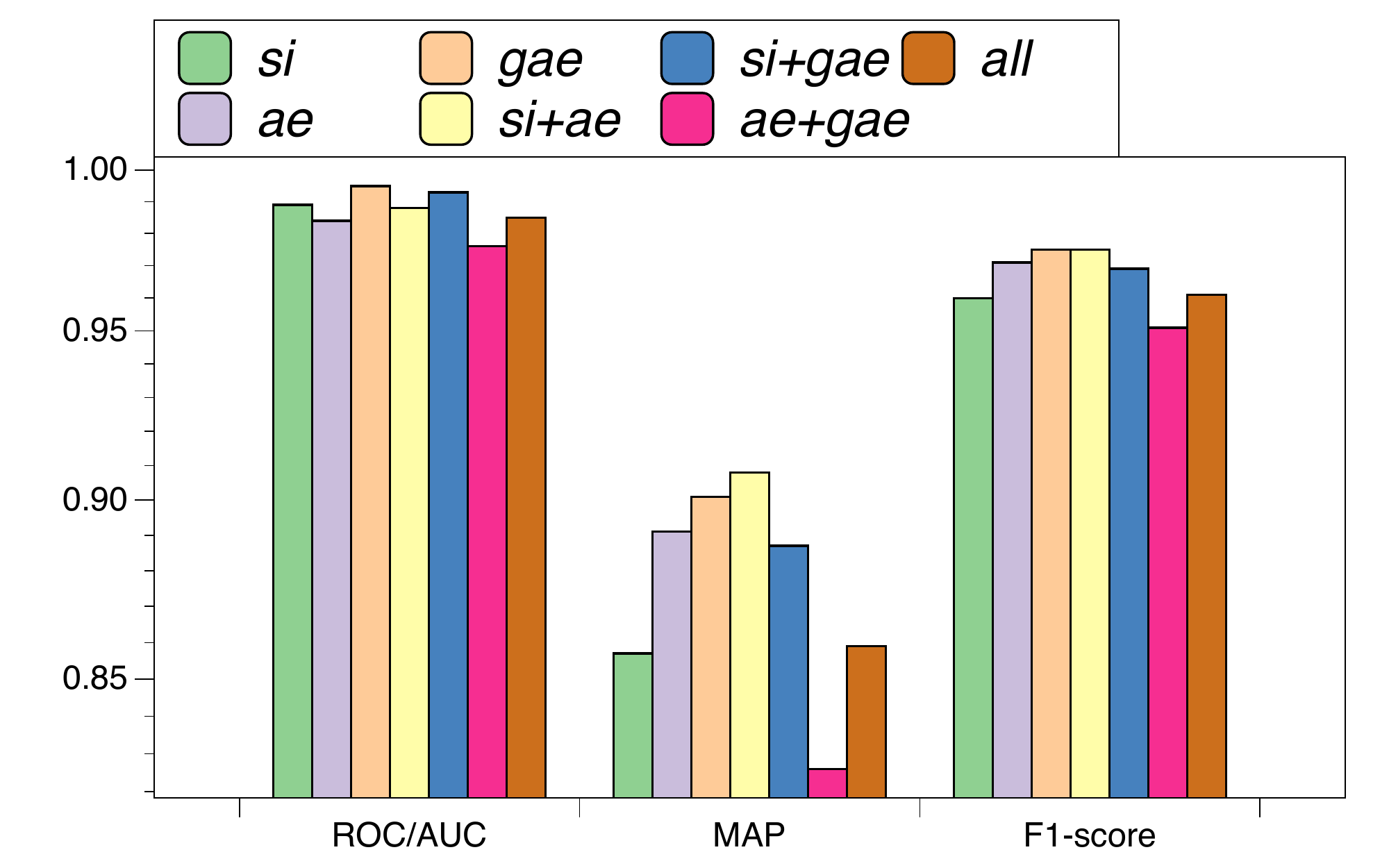}
}
\vspace{-0.35cm}
\caption{Comparison of different state representation methods.}
\label{state_study}
\vspace{-0.45cm}
\end{figure*}

\begin{figure*}[!h]
\centering
\subfigure[SVMGuide3]{
\includegraphics[width=5.2cm]{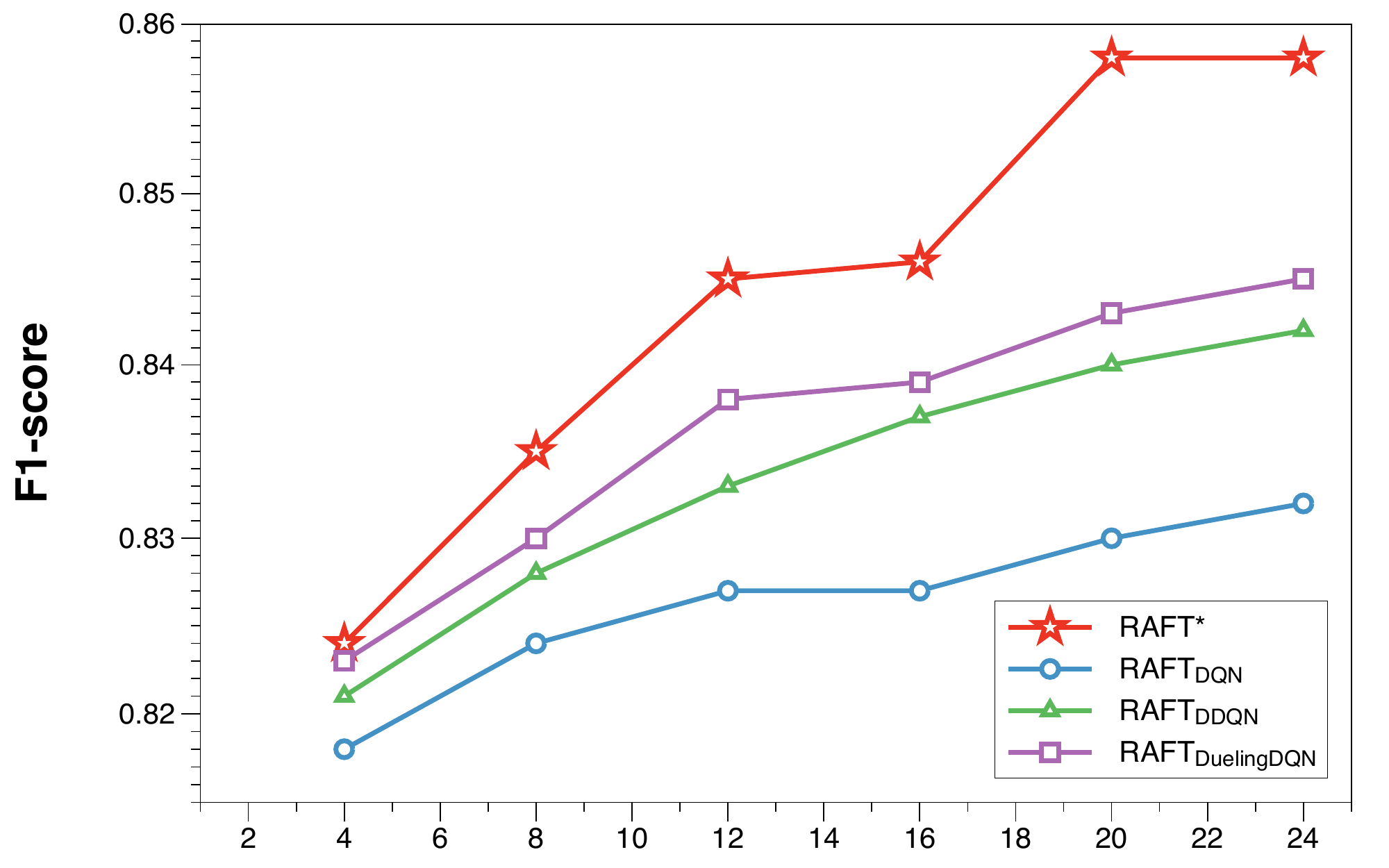}
}
\hspace{-3mm}
\subfigure[OpenML\_586]{ 
\includegraphics[width=5.2cm]{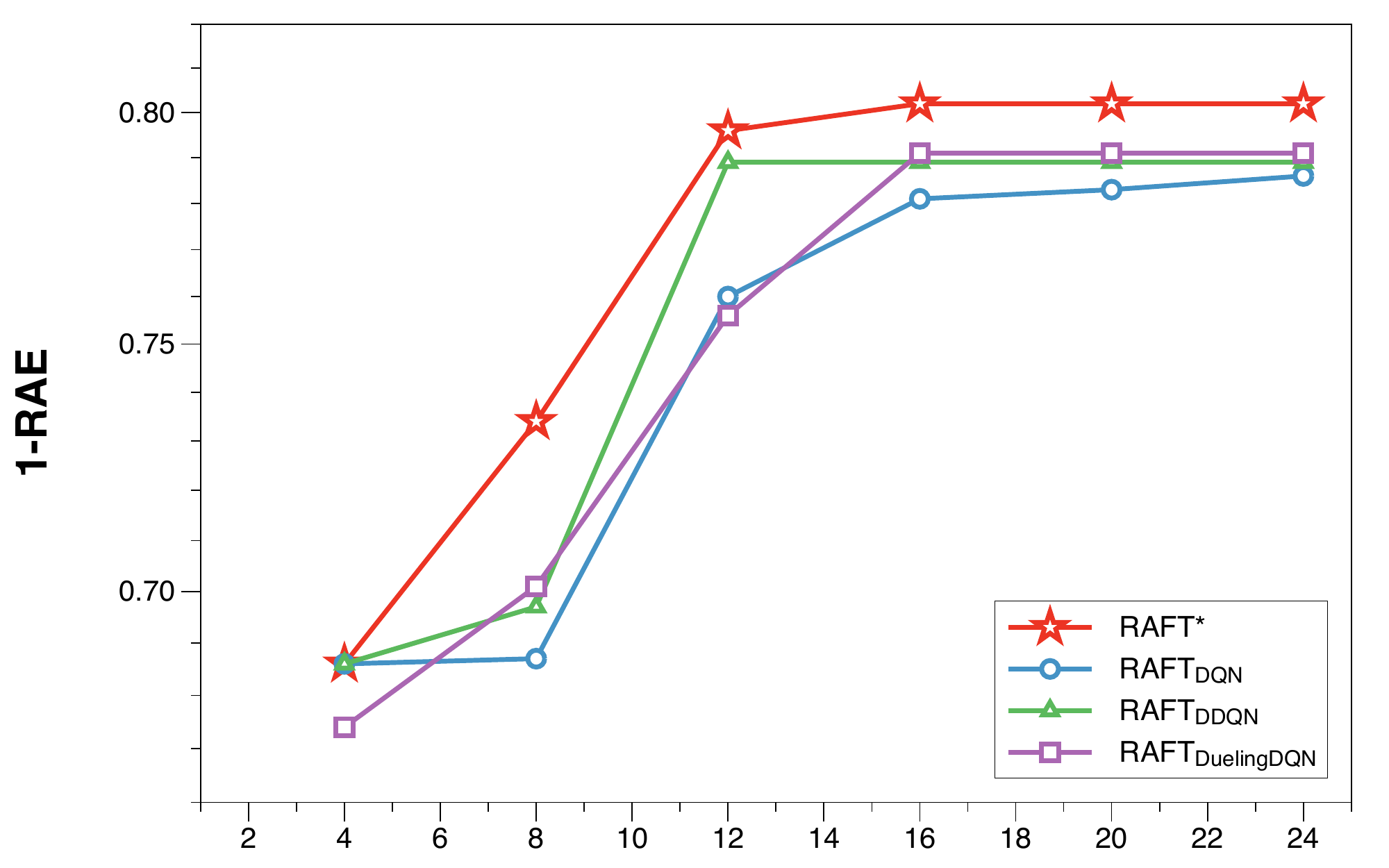}
}
\hspace{-3mm}
\subfigure[Thyroid]{
\includegraphics[width=5.2cm]{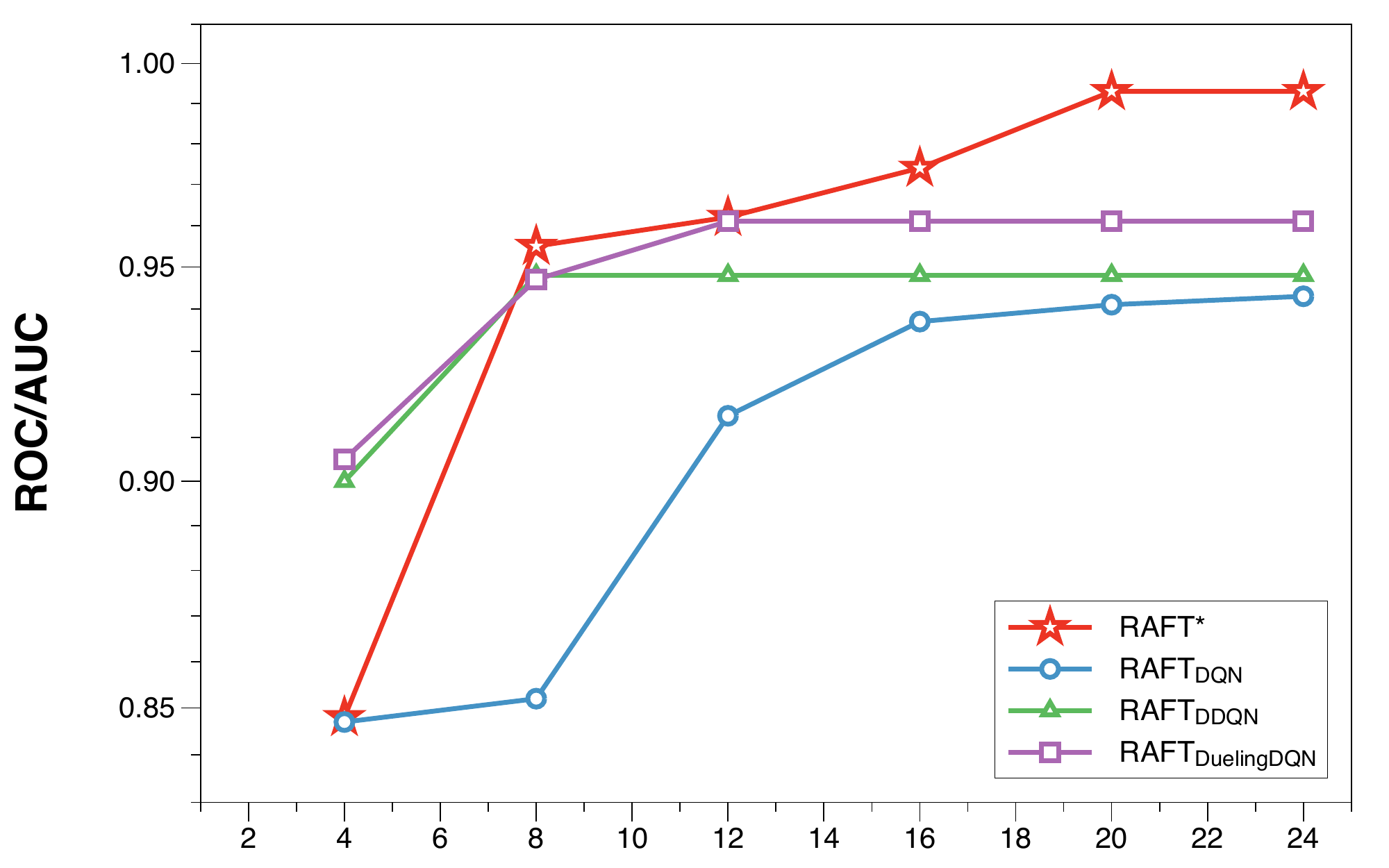}
}
\vspace{-0.35cm}
\caption{Model converge with other reinforcement learning method as backbone.}
\label{method_study}
\vspace{-0.7cm}
\end{figure*}
\vspace{-0.3cm}
\subsection{RQ1: Overall Performance}
This experiment aims to answer: 
\textit{Can RAFT effectively improve the quality of the original feature space?} 
Table~\ref{table_overall_perf} shows the overall performance of all models on all datasets. 
We can observe that RAFT significantly outperforms other baselines.
This observation indicates the effectiveness of our work in feature space reconstruction.
Another interesting observation is that RAFT beats RFG in most cases.
This observation validates that reinforced agents can model feature engineering knowledge to learn better transformation policies than random generation strategies.
We also can find that RAFT is superior to non-group-wise feature generation frameworks (\textit{i.e.,} NFS and TTG).
The underlying driver is that group-wise feature generation can efficiently refine the feature space and provide strong reward signals for reinforced agents to learn more intelligent policies.
Moreover, the superiority of RAFT compared with GRFG indicates that the actor-critic training strategy can learn more robust and effective policies than DQN-based agents.

\vspace{-0.3cm}
\subsection{RQ2: Study of the Distance Function}
This experiment aims to answer: \textit{How do different distance functions affect the quality of reconstructed feature space?} 
We adopted euclidean distance and cosine distance in the feature clustering component to observe the difference in model performance.
Figure~\ref{dis_study} shows performance comparison on different datasets.
We can find that euclidean distance outperforms cosine distance on the datasets with high standard deviation. But, the observation is the opposite on low standard deviation datasets.
A possible reason is that the value range of cosine distance is $[-1, 1]$ but the euclidean distance is $[-\text{infinity}, +\text{infinity}]$. Thus, the euclidean distance may enlarge more distances between feature groups when confronted with a high standard deviation dataset. 
It will produce more informative features to refine the feature space.
Thus, this experiment provides a strategy to customize feature distance for different datasets.

\subsection{RQ3: Study of the State Representation Methods}
\vspace{-0.4cm}
This experiment aims to answer: \textit{How do different state representation approaches affect the reconstructed feature space?} Apart from the introduced state representation methods (i.e., \textit{si}, \textit{ae}, and \textit{gae}), we also try combinations of them such as \textit{si+ae}, \textit{si+gae}, \textit{ae+gae}, and \textit{all}. 
For each combination method, we concatenated the state representations from different approaches.
 Figure~\ref{state_study} shows the comparison results.
We can notice that \textit{gae} outperforms other methods in most tasks. 
A possible reason is that \textit{gae} captures not only the knowledge of the feature set but also feature-feature correlations. 
It preserves more knowledge of the feature set in the state, which makes agents can learn effective policies better.
Another interesting observation is that although the combination-based method captures more characteristics of the feature space, their performances still cannot outperform others.
A potential interpretation is that directly concatenating different states may include redundant and noisy information, leading reinforced agents to learn suboptimal policies.

\subsection{RQ4: Comparison with Value-based Approaches}
\vspace{-0.4cm}
This experiment aims to answer: \textit{How does the training strategy affect the quality of the refined feature space?
} 
We developed three model variants of RAFT: RAFT$_\text{DQN}$, RAFT$_\text{DDQN}$, and RAFT$_\text{DuelingDQN}$ by replacing actor-critic agents with Deep Q-Network (DQN), Double DQN (DDQN), and Dueling DQN. 
 Figure~\ref{method_study} shows the comparison results. 
 We can observe that RAFT has a comparable converge efficiency in comparison to  $\text{RAFT}_{\text{DDQN}}$ and $\text{RAFT}_{\text{DuelingDQN}}$. 
 Moreover, RAFT significantly outperforms other model variants. 
 A possible reason is that actor-critic agents directly optimize the transformation policies.
 Thus, they extensively explore the high-dimensional feature space transformation tasks compared with other baselines.

\vspace{-0.3cm}

\subsection{RQ5: The Traceability of Automatic Feature Generation}
\begin{figure}[!h]
\centering
\subfigure[The Original Feature]{
\includegraphics[width=0.42\linewidth]{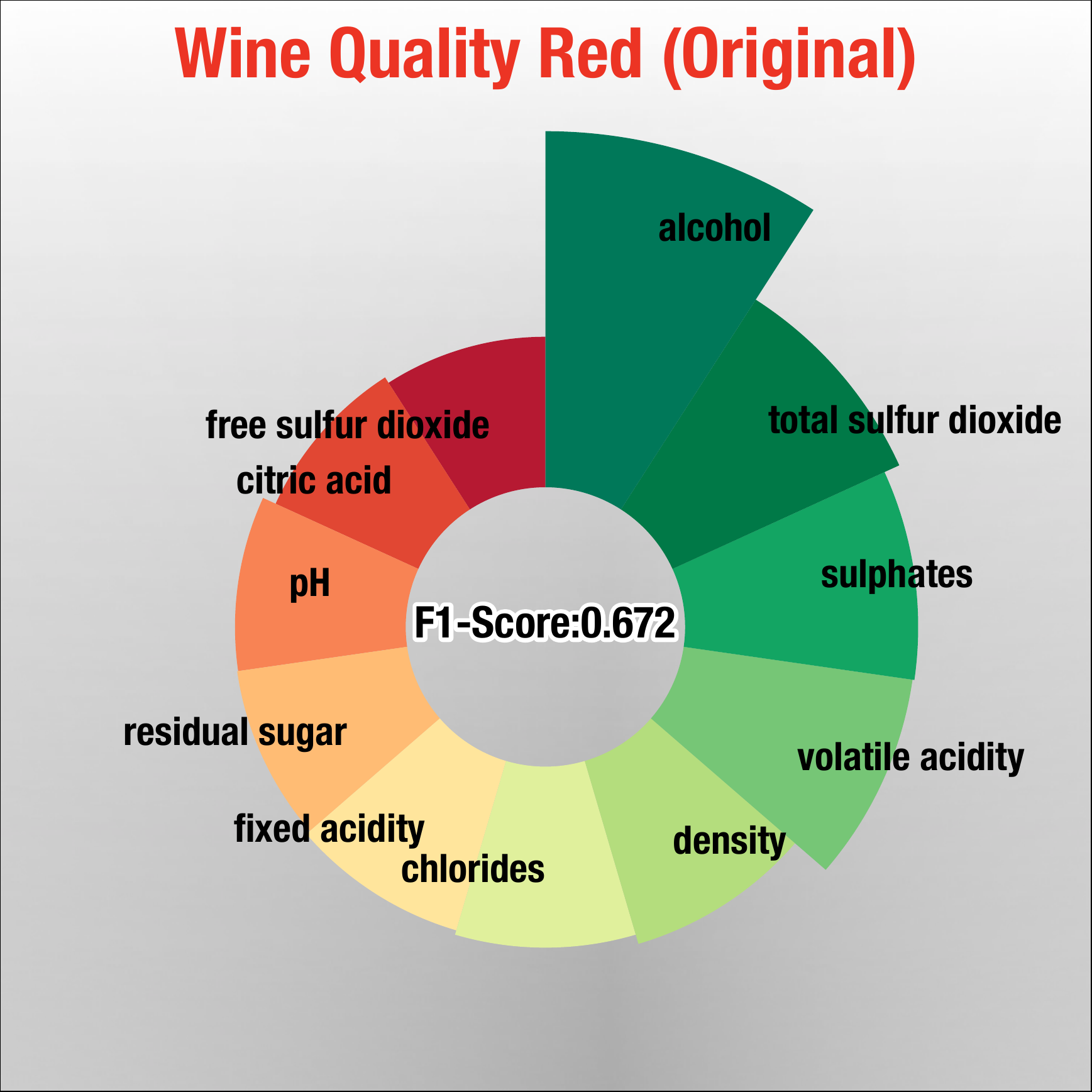}
}
\subfigure[The Generated Feature]{ 
\includegraphics[width=0.42\linewidth]{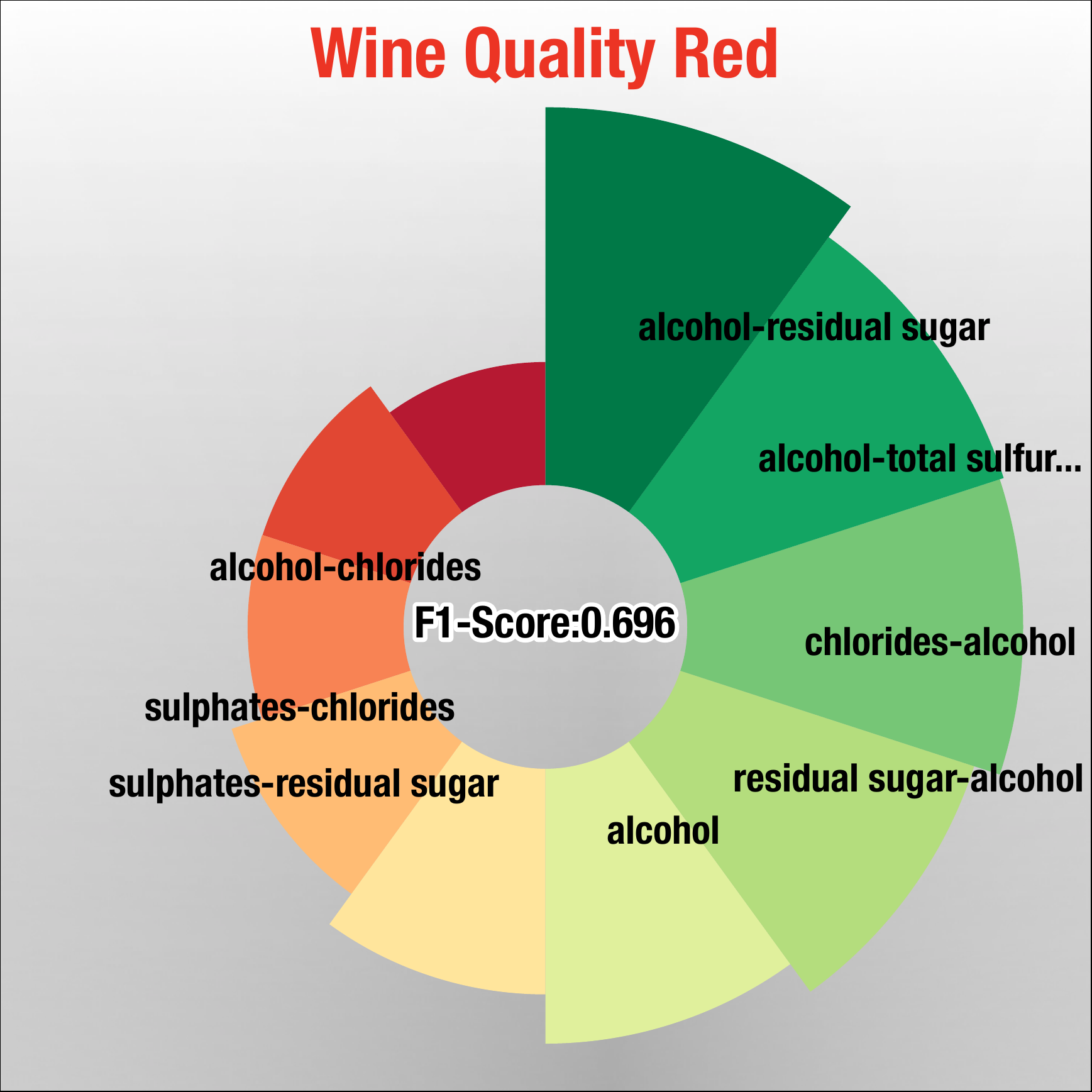}
}
\vspace{-0.3cm}
\caption{The illustration of model traceability.}
\label{traceable_study}
\vspace{-0.7cm}
\end{figure}

This experiment aims to answer: \textit{How is the traceability of the feature space generated by RAFT?}
We selected the dataset ``Wine Quality Red'' as an example to show traceability.
We visualized the original and generated features in Figure~\ref{traceable_study}. The size of each sector area represents the importance of each feature. We can find that the `alcohol' in the original dataset is far more critical than other features. However, the generated feature has a more balanced importance distribution for all features. Meanwhile, we can easily figure out the transformation process of each generated feature by its name. For instance, the most critical column in generated feature is ``alcohol$-$residual sugar'', which is generated by two original features ``alcohol'' and ``residual sugar''. 
\vspace{-0.3cm}

\section{Related Works}
\noindent\textbf{Reinforcement Learning (RL)} is the study of how intelligent agents should act in a given environment in order to maximize the expectation of cumulative rewards~\cite{sutton2018reinforcement}.
According to the learned policy, we may classify reinforcement learning algorithms into two categories: value-based and policy-based.
Value-based algorithms (\textit{e.g.} DQN~\cite{mnih2013playing}, Double DQN~\cite{van2016deep}) estimate the value of the state or state-action pair for action selection.
Policy-based algorithms (\textit{e.g.} PG~\cite{sutton2000policy}) learn a probability distribution to map state to action for action selection.
Additionally, an actor-critic reinforcement learning framework is proposed to  incorporate the advantages of value-based and policy-based algorithms~\cite{schulman2017proximal}.
In recent years, RL has been applied to many domains (e.g. spatial-temporal data mining, recommended systems) and achieves great achievements~\cite{wang2022reinforced,wang2022multi}.
In this paper, we adopted actor-critic based method to construct the cascading agents.

\noindent\textbf{Automated Feature Engineering} aims to enhance the feature space through feature generation and feature selection in order to improve the performance of machine learning models~\cite{chen2021techniques}.
Feature selection is to remove redundant features and retain important ones, whereas feature generation is to create and add meaningful variables. 
\ul{\textit{Feature Selection}} approaches include:
(i) filter methods (\textit{e.g}., univariate selection \cite{forman2003extensive}, correlation based selection \cite{yu2003feature}), in which features are ranked by a specific score like redundancy, relevance;  (ii) wrapper methods (\textit{e.g.}, Reinforcement Learning~\cite{ liu2021efficient}, Branch and Bound~\cite{ kohavi1997wrappers}), in which the optimized feature subset is identified by a search strategy under a predictive task;  (iii) embedded methods (\textit{e.g.}, LASSO \cite{tibshirani1996regression}, decision tree \cite{sugumaran2007feature}), in which selection is part of the optimization objective of a predictive task. 
\ul{\textit{Feature Generation}} methods include: (i) latent representation learning based methods, e.g. deep factorization machine~\cite{guo2017deepfm}, deep representation learning~\cite{bengio2013representation}. 
Due to the latent feature space generated by these methods, it is hard to trace and explain the extraction process.
(ii) feature transformation based methods, which use column-wise arithmetic operations~\cite{khurana2018feature,chen2019neural} or group-wise arithmetic operations ~\cite{wang2022group, xiao2022self} to generate new features. 
\vspace{-0.3cm}

\section{Conclusion}
In this paper, we propose a traceable automatic feature transformation framework called RAFT. The RAFT can utilize cascading actor-critic agents to develop optimal features, hence enhancing the performance of subsequent tasks.
We design an FG-cluster algorithm with two distance functions based on a group-wise feature generation procedure for greater efficiency.
In addition, we offer three feature state representation approaches to assist cascade agents in evaluating the current feature set and, as a result, making more informed decisions.
Extensive studies are conducted on RAFT to demonstrate the efficacy of each component and its application potential in numerous research fields. 
\vspace{-0.3cm}
\vspace{-0.1cm}
\section{Acknowledgement}
This work is partially supported by IIS-2152030, IIS-2045567, and IIS-2006889.
\vspace{-0.4cm}
\bibliographystyle{siam}

\bibliography{z.meng, z.acm, z.yanjie}
\end{document}